\documentclass{article}
\PassOptionsToPackage{numbers,sort&compress}{natbib}
\usepackage[preprint]{neurips_2026}
\usepackage[utf8]{inputenc} 
\usepackage[T1]{fontenc}    
\usepackage{hyperref}       
\usepackage{url}            
\usepackage{booktabs}       
\usepackage{amsfonts}       
\usepackage{nicefrac}       
\usepackage{microtype}      
\usepackage{xcolor}         
\usepackage{graphicx}
\usepackage{amsmath,amssymb}
\usepackage{tabularx,array,multirow}

\title{Deep Reinforcement Learning for Dynamic Origin-Destination Matrix Estimation in Microscopic Traffic Simulations Considering Credit Assignment}

\author{%
  Donggyu Min \\
  Seoul National University \\
  Seoul, 08826 \\
  \texttt{dgmin@snu.ac.kr}
  \And
  Seongjin Choi \\
  University of Minnesota \\
  Minneapolis, MN 55455 \\
  \texttt{chois@umn.edu}
  \And
  Dong-Kyu Kim\thanks{Corresponding author.} \\
  Seoul National University \\
  Seoul, 08826 \\
  \texttt{dongkyukim@snu.ac.kr}
}

\begin{document}
\maketitle

\begin{abstract}
  This paper focuses on dynamic origin-destination matrix estimation (DODE), a crucial calibration process necessary for the effective application of microscopic traffic simulations. The fundamental challenge of the DODE problem in microscopic simulations stems from the complex temporal dynamics and inherent uncertainty of individual vehicle dynamics. This makes it highly challenging to precisely determine which vehicle traverses which link at any given moment, resulting in intricate and often ambiguous relationships between origin-destination (OD) matrices and their contributions to resultant link flows. This phenomenon constitutes the credit assignment problem, a central challenge addressed in this study. We formulate the DODE problem as a Markov Decision Process (MDP) and propose a novel framework that applies model-free deep reinforcement learning (DRL). Within our proposed framework, the agent learns an optimal policy to sequentially generate OD matrices, refining its strategy through direct interaction with the simulation environment. This approach was evaluated through a toy experiment on the Nguyen–Dupuis network and a case study utilizing an actual highway subnetwork spanning Santa Clara and San Jose. Experimental results show that the proposed method consistently improves calibration performance relative to the strongest conventional baseline, reducing link-flow MSE by 23.7\% in the toy experiment and by 59.2–88.3\% in the real-world case study. By reframing DODE as a sequential decision-making problem, our approach addresses the credit assignment challenge through a learned policy and provides a novel framework for calibration of microscopic traffic simulations.
\end{abstract}



\section{Introduction}

Advances in data acquisition and computational power have enabled sophisticated, computationally intensive transportation strategies \citep{ref1}. Modern strategies, such as speed harmonization, lane-changing control, and adaptive traffic signal control, are now predominantly designed and evaluated at the microscopic level, marking a significant shift in traffic management paradigms \citep{ref2, ref3, ref4}. This emphasis on microscopic analysis underscores the importance of microscopic traffic simulations.

The reliability of microscopic traffic simulation depends on a rigorous calibration process that replicates real-world traffic dynamics \citep{ref5, ref6, ref7}. This process typically involves minimizing a loss function that quantifies the discrepancy between simulation outputs and field observations \citep{ref8}. A critical component of calibration is the dynamic origin-destination matrix estimation (DODE) problem, which aims to adjust traffic demand inputs to reproduce observed data, such as link flows \citep{ref9}.

A conventional approach to the DODE problem involves formulating a bi-level optimization problem \citep{ref10, ref11, ref12}. This consists of an upper-level problem to determine the optimal origin-destination (OD) matrix and a lower-level problem, typically a dynamic assignment problem, that maps the OD matrix to link flows. While this bi-level approach offers computational tractability and modeling flexibility \citep{ref11}, applying it to calibrate microscopic simulations presents considerable challenges.

The primary challenge stems from the nature of microscopic dynamics, where the impacts of sequential OD matrices on link flows are not independent across time but are temporally complex and highly stochastic. In microscopic simulations, dynamics are modeled at the individual vehicle level. This granularity, combined with complex driver interactions, makes it infeasible to deterministically trace future vehicle trajectories \citep{ref9, ref13, ref14, ref15}. These interactions induce unpredictable network loading patterns, which in turn make it difficult to attribute observed link flows to the specific, time-lagged OD inputs—a fundamental issue known as the credit assignment problem \citep{ref16, ref17}.

To address this credit assignment problem, we reformulate the DODE problem as a sequential decision-making problem using a Markov Decision Process (MDP) framework. The MDP explicitly accounts for long-term stochastic impacts of sequential OD matrices, offering a structured approach to the credit assignment issue that conventional optimization methods often overlook.

However, solving the DODE problem as an MDP introduces two technical challenges. The first is the combination of high-dimensional states and intractable transition dynamics, which renders traditional MDP solvers, such as dynamic programming, computationally infeasible due to the curse of dimensionality and the lack of an explicit transition model \citep{ref18}. To overcome this issue, we employ a deep reinforcement learning (DRL) approach, which enables an agent to learn an optimal policy through direct interaction with the simulation environment.

The second challenge is the instability of policy training due to a large action space, stochastic noise, and delayed rewards \citep{ref19}. To address this, a specialized DRL approach is required. A large, combinatorial action space complicates exploration, while stochastic noise and delayed rewards can lead to unstable policy gradients and slow convergence. We employ proximal policy optimization (PPO) \citep{ref20} with a multi-binary action parameterization. Structuring the action space as multi-binary reduces combinatorial complexity by letting the agent decide whether to release demand for each OD pair at each interval. In addition, PPO's clipped surrogate objective implicitly limits the update magnitude, mitigating instability induced by stochasticity. PPO's actor-critic architecture, combined with generalized advantage estimation (GAE) \citep{ref21}, reduces variance and facilitates effective long-horizon credit assignment by integrating immediate rewards with value function estimates. The main contributions of this paper are as follows:
\begin{enumerate}
    \item We frame the challenge of DODE in microscopic traffic simulation as a \textbf{\textit{credit assignment problem}} and propose an \textbf{\textit{MDP framework}} to formalize it.
    \item We propose a \textbf{\textit{model-free DRL approach}} to handle high-dimensional observations and intractable transition dynamics inherent in this MDP formulation.
    \item We employ a \textbf{\textit{multi-binary PPO}} to manage the combinatorial action space and achieve robust convergence under stochasticity and delayed rewards.
\end{enumerate}

The remainder of this paper is organized as follows. The next section reviews the relevant literature on the DODE problem. We then present our problem formulation designed to address the identified research gap, followed by the proposed methodology to solve it. Subsequently, we present the experimental results and discuss their implications. The final section concludes the paper and suggests directions for future research.

\section{Literature Review}

\subsection{Conventional Approaches to Dynamic Origin-Destination Matrix Estimation}

The estimation of OD matrices has traditionally been studied in both planning and operational contexts. In the planning stage, researchers estimate travel demand for a given region to predict unknown travel patterns. In contrast, the operational stage aims to calibrate demand inputs to align outputs from traffic models, such as simulations, with observed ground-truth data. The focus of this study is on the operational stage. Specifically, we aim to identify the optimal input trajectory of OD matrices for a microscopic model, such that the generated link flows closely align with observed ground-truth data.

Initial research on estimating OD matrices from observed link flows established foundational concepts and methodologies that informed subsequent developments \citep{ref22}. To reproduce observed link flows requires distributing trips defined by the OD matrix onto specific routes within the network. Initially, this problem was defined using a time-independent assignment matrix to relate the OD matrix to link flows and formulated as a single-objective optimization problem \citep{ref23}. As research advanced, the DODE problem emerged, aiming to capture realistic driver behavior by dividing the analysis period into shorter intervals and dynamically distributing time-dependent OD matrices along network paths. However, using a fixed assignment matrix proved inadequate for accurately representing the dynamic and temporally varying relationship between OD matrices and observed link flows.

To address this issue, researchers expanded traditional formulations. Cremer and Keller \citep{ref24} formulated the DODE problem as a constrained optimization problem aiming to minimize the squared errors between estimated and observed link flows to determine time-dependent OD matrices. Cascetta \textit{et al.} \citep{ref25} treated the problem as an optimization problem based on dynamic traffic assignment models, using time-dependent route choice proportions to approximate the relationship between OD matrices and observed link flows. Ashok and Ben-Akiva \citep{ref26} applied a state-space model to estimate OD matrices by modeling their deviations from historical data. Bierlaire and Crittin \citep{ref27} formulated the DODE problem as a single optimization problem and compared the performance of the Kalman filter with the more efficient LSQR algorithm for its solution. While these methods are still occasionally revisited and extended for their computational convenience \citep{ref28}, their simplifying treatment of congestion and route-choice feedback can be restrictive when calibrating congested networks and high-resolution simulations. This motivates a bi-level optimization framework that explicitly integrates OD matrix estimation with dynamic traffic assignment.

\subsection{The Bi-level Optimization Framework}

An alternative approach to solving the DODE problem is to formulate it as a bi-level optimization problem. To handle real-world congestion effects and measurement errors, Yang \textit{et al.} \citep{ref10} argued that OD matrix estimation should be integrated with equilibrium traffic assignment, rather than being treated separately. This formulation is based on the concept that upper-level decisions influence the lower-level problem, whose outcomes subsequently inform the upper-level evaluation \citep{ref29}. The upper-level problem aims to find a sequence of OD matrices that minimizes discrepancies between observed and simulated data, whereas the lower-level problem is the DTA problem, which involves assigning trips based on the equilibrium travel times experienced by users \citep{ref9, ref30}.

The bi-level optimization formulation is recognized as a promising framework for addressing the DODE problem \citep{ref11}. In contrast to methods seeking a mutually consistent solution, the bi-level formulation models the lower-level's optimal response within the upper-level's decision-making. Its anticipatory nature guides the system toward a more globally optimal state, yielding better solutions \citep{ref31}. This decomposition also provides flexibility by allowing specialized techniques for each level.

Various approaches based on this formulation have emerged. For the upper-level problem, algorithms such as simultaneous perturbation stochastic approximation (SPSA) and its variations, as well as Bayesian optimization (BO) and its variants, have been developed \citep{ref9, ref32}. Meanwhile, advances in computational technology have enabled widespread use of traffic simulations to capture realistic traffic dynamics in solving the lower-level problem \citep{ref14, ref30}. Furthermore, researchers have developed more accurate and efficient DODE methods by leveraging additional data sources \citep{ref12, ref33, ref34, ref35}, advancing algorithmic sophistication \citep{ref36}, and employing surrogate models \citep{ref7, ref37, ref38}. 

Recent work has further focused on improving sample efficiency and scalability for high-dimensional simulation-based DODE by leveraging spatiotemporal structure, dimensionality reduction, and surrogate-assisted evaluation. For example, Peng \textit{et al.} propose an efficient framework that integrates a spatiotemporal pattern-based dimensionality reduction step with metamodel-based acceleration to address high dimensionality and computational inefficiency \citep{ref39}. In a complementary direction, Tang \textit{et al.} leverage potential spatiotemporal information in dynamic OD matrices to guide calibration updates and incorporate macroscopic traffic relationships in the objective to better capture nonlinear network dynamics \citep{ref36}. Related simulation-based transportation optimization studies also emphasize the importance of exploration strategies; Tay and Osorio propose a problem-informed sampling mechanism derived from an analytical network model to improve search efficiency in high-dimensional settings under tight simulation budgets \citep{ref40}. Finally, benchmark efforts such as BO4Mob underscore that OD matrix estimation in realistic networks constitutes a high-dimensional, stochastic, and non-differentiable black-box optimization problem, motivating systematic evaluation of scalable algorithms \citep{ref41}. Comprehensive reviews of recent approaches can be found in the papers of Osorio \citep{ref37} and Huo \textit{et al.} \citep{ref9}.

\subsection{Research Gap and Proposed Direction}

Despite the advancements, applying these conventional frameworks to calibrate microscopic simulations reveals a fundamental challenge. As previously noted, the temporally complex and stochastic relationships between sequential OD matrices and resulting link flows create a significant credit assignment problem. Existing methods based on a bi-level framework are categorized into two approaches: simultaneous and sequential optimization \citep{ref35}. Simultaneous optimization handles temporal complexity and stochasticity implicitly but suffers from an immense computational burden as the dimension of the decision variables increases. Conversely, sequential optimization reduces complexity but is often myopic, optimizing within finite horizons and failing to capture the long-term consequences of decisions. This creates a critical trade-off between computational complexity and analytical myopia, limiting the effectiveness of current solutions for the credit assignment problem.

Notably, most recent advances primarily aim to alleviate the computational burden of optimization (e.g., via dimension reduction, structured gradient approximation, and improved exploration), but they remain within an optimization-based calibration paradigm that directly searches over a pre-parameterized OD demand trajectory, without casting temporal credit assignment under delayed and stochastic propagation as a sequential decision problem. Accordingly, while these methods can substantially improve sample efficiency, the complexity–myopia trade-off persists when calibrating microscopic simulations at finer temporal resolutions, where delayed and downstream effects are critical.

To bridge this research gap, we reframe the DODE problem as a sequential decision-making task, formally structured within the MDP formulation. This perspective partially aligns with a recent conceptual shift that reformulates DODE as an optimal control problem \citep{ref42}. While partially observable formulations (POMDPs) could be considered to model unmeasured microscopic states, they require belief-state estimation and substantially increase computational complexity in learning and planning \citep{ref43, ref44}. In this study, we adopt a standard MDP with carefully designed state augmentation and short input timesteps, which we find sufficient to capture long-horizon effects while remaining computationally tractable. This approach aligns with a growing body of research that successfully applies standard MDPs to complex, stochastic transportation problems \citep{ref45}.

Conceptually, the MDP framework synthesizes the strengths of both prior approaches. It decomposes the problem into a series of decisions over time, like sequential optimization, thereby avoiding the exponential growth in complexity that simultaneous methods face. However, unlike myopic sequential methods, the MDP's objective is to maximize a long-term, cumulative reward. This formulation inherently forces the model to account for the delayed, downstream consequences of current actions, thus formally addressing the credit assignment problem and overcoming analytical myopia. The MDP, therefore, provides a theoretically sound pathway to achieving a long-term optimal perspective with the tractability of a sequential process.

However, while the MDP offers a robust conceptual solution, its practical implementation is challenged by microscopic simulations. The state space is extraordinarily high-dimensional—a classic challenge known as the curse of dimensionality—and the state transition dynamics are intractable and stochastic \citep{ref46}. Consequently, traditional MDP solvers, such as dynamic programming, are computationally infeasible. To bridge this gap between theory and practice, we employ a model-free DRL approach. DRL has emerged as the state-of-the-art methodology for solving complex MDPs precisely because it uses deep neural networks as function approximators to handle high-dimensional states and can learn effective policies from direct interaction without an explicit transition model \citep{ref45}. DRL thus provides the necessary tools to solve the formulated MDP, enabling an agent to learn an optimal policy within the complex environment.

By integrating the MDP framework with a model-free DRL approach, we demonstrate a practical path to overcoming the trade-off between complexity and myopia. This synthesis establishes a novel paradigm, enabling a previously intractable level of granularity in demand estimation and opening new avenues for high-fidelity model calibration and control.

\section{Methods}

\subsection{Key Concept}

This study estimates sequences of OD matrices for the offline calibration of microscopic traffic simulations. The primary objective is to simulate detector data as link flows aggregated over 5-minute intervals, ensuring they closely match ground-truth observations. A fundamental challenge arises from the temporally complex and stochastic relationships inherent in vehicle-level dynamics.

\begin{figure}[htbp]
  \centering
  \includegraphics[width=1.0\linewidth]{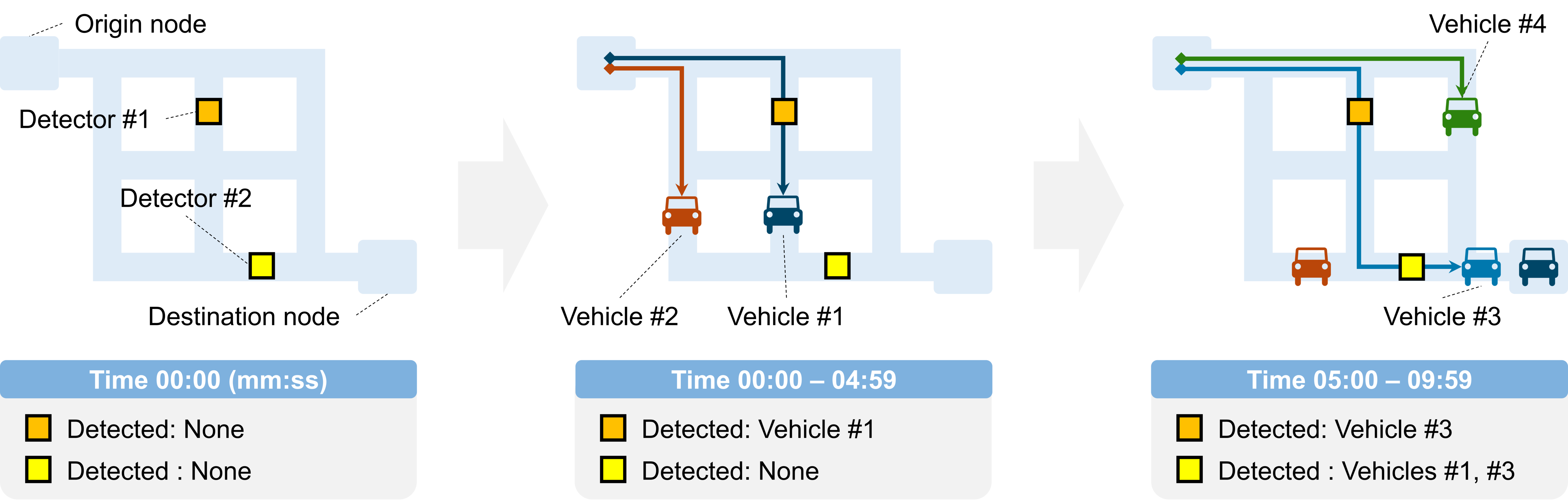}
  \caption{Comparison of detector data for vehicles with the same OD pair}
  \label{fig1}
\end{figure}

The input-output relation is illustrated in Figure~\ref{fig1}, which shows how vehicles with the same OD pair can influence detector data differently. For instance, vehicles \#1 and \#2 depart simultaneously but take different paths, resulting in their detection at separate detectors. Conversely, vehicles \#1 and \#3 depart at different timesteps yet are recorded within the same timestep due to variable network conditions, such as congestion. Any detector may not capture vehicle \#4 if its chosen trajectory does not traverse a monitored link, or if it exits the network or remains in transit after the simulation. These complex and stochastic outcomes make it analytically intractable to attribute observed link flows back to specific departure decisions, which constitutes the core of the credit assignment problem \citep{ref19}.

To address this challenge, we reformulate the DODE problem as a sequential decision-making task. This task is formally structured using an MDP framework \citep{ref19}. We represent the network dynamics, driven by vehicle departures and their subsequent travel, as a discrete-time system with sufficiently short timesteps. This discrete representation allows the Markov property, which posits that the subsequent state depends only on the current state and the chosen action, to serve as a reasonable approximation of real-world traffic dynamics. By formulating the DODE process as an MDP with stochastic transitions, we represent the temporally complex and stochastic evolution induced by microscopic interactions. The transition kernel is not available in closed form in a microscopic simulator; instead, the simulator provides samples of the transition kernel.

\begin{figure}[htbp]
  \centering
  \includegraphics[width=1.0\linewidth]{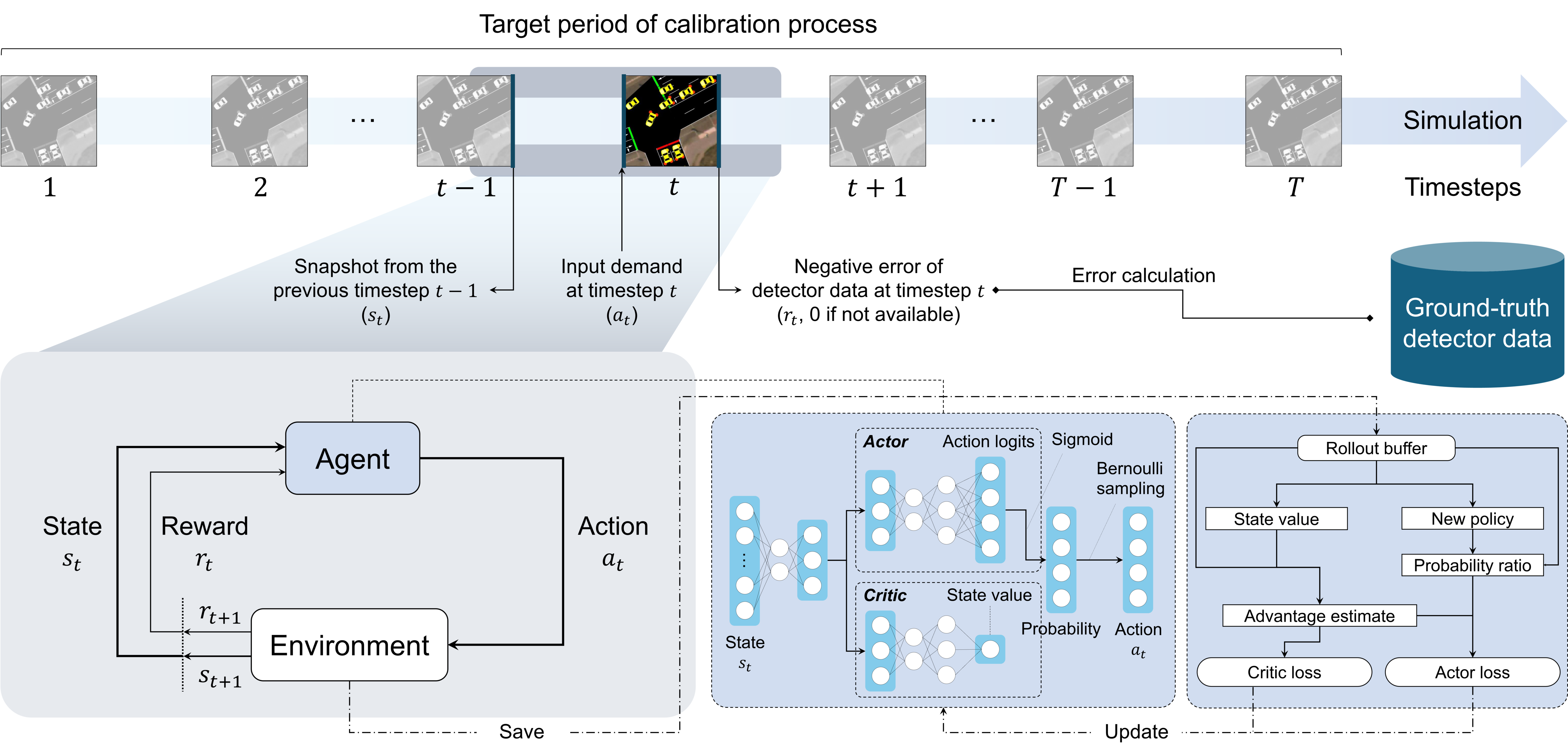}
  \caption{Overall framework of the proposed method}
  \label{fig2}
\end{figure}

Although the MDP provides a suitable theoretical framework, its direct solution is computationally intractable due to the high-dimensional state space. We therefore employ a model-free DRL approach, allowing an agent to learn an optimal policy through direct interaction with the simulation environment, as shown in Figure~\ref{fig2}. A policy maps the current network state to a dispatch decision for each OD pair. The agent's learning is driven by maximizing the long-term cumulative reward, defined as the negative error between simulated and ground-truth data. The policy with the best performance in reproducing the ground-truth data during training is selected, and the demand input trajectory it generates constitutes our final output. This trajectory consists of a sequence of binary departure decisions for each OD pair at every timestep.

The subsequent sections detail the specific MDP formulation and the DRL algorithm employed. To promote and enable transparent and reproducible research, codes and data from this study are shared with the community (refer to the \textbf{Data Availability}).

\subsection{Problem Statement}

\subsubsection{Notations}

Before presenting the mathematical formulation, we define the key notations used throughout this section. For clarity of reference, notations are summarized in Table~\ref{tab:t1}.


\begin{table}[!t]
  \centering
  \caption{Notations}
  \label{tab:t1}
  \setlength{\tabcolsep}{4pt}
  \renewcommand{\arraystretch}{1.15}
  \begin{tabularx}{\linewidth}{@{}>{\centering\arraybackslash}p{0.18\linewidth}X@{}}
    \toprule

    \multicolumn{2}{@{}l}{\textit{Time}} \\
    \midrule
    $\mathcal{T}$ 
    & Set of input time index $(=\{1,2,\ldots,T\})$ \\
    $T$ 
    & Maximum index of inputs \\

    \midrule
    \multicolumn{2}{@{}l}{\textit{Action}} \\
    \midrule
    $\mathcal{A}$ 
    & Action space \\
    $a$ 
    & Input trajectory covering the entire period 
    $(=\{a_t\}_{t=1}^{T},\ a_t \in \mathcal{A})$ \\
    $a^{*}$ 
    & Optimal input trajectory \\
    $a_t$ 
    & OD matrix at $t$ \\

    \midrule
    \multicolumn{2}{@{}l}{\textit{State}} \\
    \midrule
    $\mathcal{S}$ 
    & State space \\
    $s$ 
    & State trajectory for all timesteps 
    $(=\{s_t\}_{t=0}^{T},\ s_t \in \mathcal{S})$ \\
    $s_t$ 
    & Traffic network and context state at $t$ \\
    $s_0$ 
    & Initial state before any input \\
    $N_l$ 
    & Number of vehicles for link $l$ \\
    $\bar{v}_l$ 
    & Average speed for link $l$ \\
    $v_f$ 
    & Free-flow speed \\

    \midrule
    \multicolumn{2}{@{}l}{\textit{Detector data}} \\
    \midrule
    $\mathcal{K}$ 
    & Set of output time index $(=\{1,2,\ldots,K\})$ \\
    $K$ 
    & Maximum index of outputs \\
    $d$ 
    & Ground-truth detector data covering the entire period 
    $(=\{d_k\}_{k=1}^{K})$ \\
    $d_k$ 
    & Ground-truth detector data at $k$ \\
    $d'$ 
    & Simulated detector data covering the entire period 
    $(=\{d'_k\}_{k=1}^{K})$ \\
    $d'_k$ 
    & Simulated detector data at $k$ \\

    \midrule
    \multicolumn{2}{@{}l}{\textit{Functions}} \\
    \midrule
    $\mathcal{L}$ 
    & Loss function corresponding to the calibration error \\

    $\mathcal{M}$ 
    & \begin{minipage}[t]{\linewidth}
        Mapping function, which is a rule that defines the temporal relationship 
        between $\mathcal{T}$ and $\mathcal{K}$. $\mathcal{P}(\cdot)$ denotes the power set of an arbitrary set.

        \vspace{0.3em}
        For $T>K$, 
        $\mathcal{M}: \mathcal{K} \rightarrow \mathcal{P}(\mathcal{T})$, 
        $\mathcal{M}=\psi(k)\subseteq\mathcal{T}$. 
        When the number of inputs is greater than the number of outputs, 
        the output index set is mapped to the power set of the input index set. 
        Therefore, the function returns the set of input indices $t$ that 
        contribute to output index $k$.
      \end{minipage} \\

    $\mathcal{F}$ 
    & Microscopic simulation model structured by the mapping function 
    $\mathcal{M}$ for generating the simulated detector data \\
    $p$ 
    & State transition function \\
    $r$ 
    & Reward function \\
    $\gamma$ 
    & Discount factor \\

    \bottomrule
  \end{tabularx}
\end{table}

\subsubsection{Dynamic origin-destination matrix estimation problem}
The DODE problem seeks to determine an input sequence of OD matrices that best reproduces a given ground-truth dataset. This ground-truth dataset comprises detector data, including link flows, collected at discrete timesteps. In this study, detector data refers to vectors composed of link flows collected every 5 minutes at specific links. The OD matrix $a_t$ at the timestep $t$ represents the flow rates for each OD pair trip starting at the origin node $i$ and arriving at the destination node $j$, which can be expressed as a vector,
\begin{equation}
\label{eq:eq1}
a_t=\left[a_{(1,1)}, a_{(1,2)}, \cdots, a_{(i, j)}, \cdots, a_{(\max (I), \max (J))}\right]
\end{equation}
where $a_t$ is the OD demand vector at timestep $t$, $a_{(i,\ j)}$ is the flow rate for trips with origin $i$ to destination $j$, $I$ is the set of indices of origin nodes ($i\in I$), and $J$ is the set of indices of destination nodes ($j\in J$).

This study addresses an inverse problem for input trajectory estimation in microscopic simulation models. Given a known initial state and a set of empirical detector data, the objective is to deduce the optimal input trajectory, $a^*$, that minimizes the deviation between the simulation output and the observed data. An objective function quantifies this deviation, $\mathcal{L}$, defined as the sum of squared Euclidean distances over the entire simulation horizon as follows:
\begin{equation}
\label{eq:eq2}
\min _{a \in \mathcal{A}} \mathcal{L}\left(a, s_0, d\right)=\left\|d^{\prime}-d\right\|_2^2=\sum_{k=1}^K\left\|d_k^{\prime}-d_k\right\|_2^2,
\end{equation}
subject to $d^{\prime}{ }_k \sim \mathcal{F}\left(a \mid s_0\right)$ for all $k \in \mathcal{K}$, with a given initial state $s_0 \in \mathcal{S}$.

In this paper, we formulate the DODE problem within an MDP framework, defining it as a process of sequentially determining the optimal departure decision for each OD pair given the current network state. An MDP is formally defined by a tuple $\langle\mathcal{S}, \mathcal{A}, p, r, \gamma\rangle$, comprising a state space $\mathcal{S}$, an action space $\mathcal{A}$, a state transition function $p$, a reward function $r$, and a discount factor $\gamma \in[0,1]$. This MDP-based approach allows for the optimization of long-term outcomes, thereby addressing the myopic nature of sequential methods. Furthermore, by learning a policy instead of optimizing the entire trajectory at once, it circumvents the complexity inherent in the simultaneous formulation. An empirical analysis of MDP is provided in \textbf{Appendix A}.

\subsubsection{State}
In MDP, a state is a complete description of the agent and environment at a specific point in time. In this study, a state consists of two elements: network state and context state. First, to maintain dimensional consistency, we propose a link-based network state, which is represented as a 1-dimensional vector containing the link vehicle counts $N_l$ and the link average speed $\bar{v}_l$ for each link. Link vehicle counts ($N_l$) encode the instantaneous distribution of vehicles across links, which is a direct proxy for vehicles-in-network accumulation and queue formation. Link average speeds ($\bar{v}_l$) provide a compact representation of congestion state and propagation, reflecting how current network loading affects future movements and travel times over subsequent steps. Together, ($N_l$, $\bar{v}_l$) serve as a link-based snapshot of network conditions that the simulator uses to propagate vehicles forward.

Additionally, a context state is proposed to provide the agent with contextual information about the calibration target. The context state consists of the current timestep and detector data memory. First, the timestep index provides temporal information to the agent. While it does not directly supply information about the target we need to fit, it helps the agent discover an implicit reward function through training. Second, detector data memory provides information on how much of the target has been achieved by indicating how many vehicles have been detected by each detector within a single detection interval. This memory is initialized as a zero vector whenever the aggregation of detector data is completed and updated. The state configuration helps determine how much demand to input into which OD pairs by considering the current network state, the current target value, and how much the objective has been achieved.

\begin{figure}[htbp]
  \centering
  \includegraphics[width=1.0\linewidth]{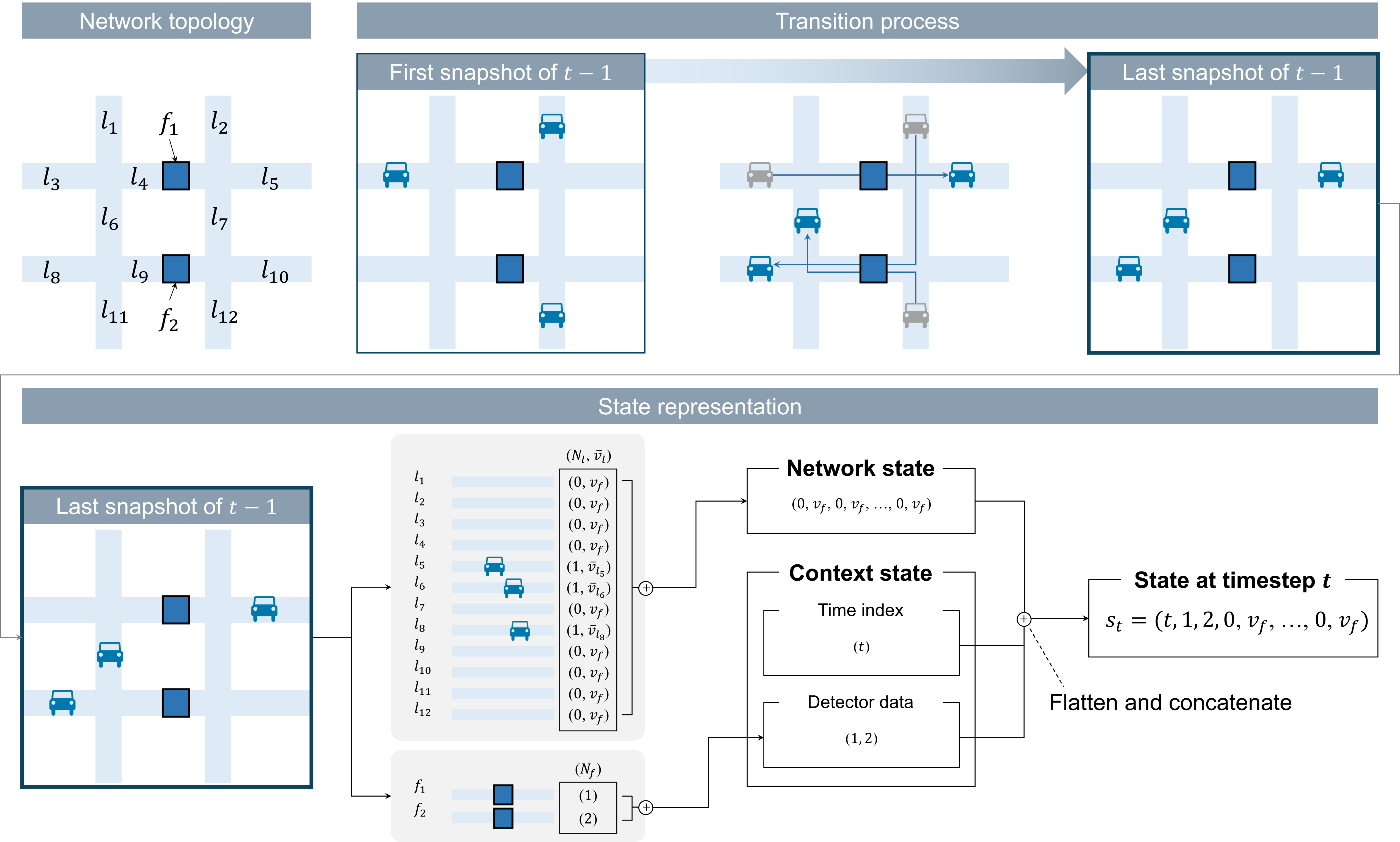}
  \caption{State description}
  \label{fig3}
\end{figure}

An example of a state is shown in Figure~\ref{fig3}. This figure explains how to construct the state $s_t$ observed at the point when the agent needs to take action $a_t$ at timestep $t$. $s_t$ depends entirely on the last snapshot in the simulation at timestep $t-1$, which is the result of vehicles moving along time-dependent shortest paths during the input interval. We extract the network state and context state from the last snapshot, then flatten and concatenate them to create $s_t$. At this point, the average speed of links with no vehicles is assumed to be the free-flow speed $v_f$. In this example, the network consists of 12 links, and two detectors are installed; therefore, the dimension of the network state is 24, and the dimension of the context state is 3. Therefore, the dimension of the state is 27. Experiments and discussions regarding the design of state are in \textbf{Appendix B}.

\subsubsection{Action}
In the proposed method, the input interval is set to a short interval of 5 seconds. This reduces the probability of unobservable events occurring between the current state and the next state, thereby improving the stability of the MDP framework. According to this experimental setup, the agent only needs to decide whether to dispatch a vehicle for each OD pair in the current step. All elements of the OD matrix $a_t$ take values of 0 or 1, allowing it to be represented as a flattened binary vector.

\subsubsection{State transition probability}
The state transition probability $p\left(s_{t+1} \mid s_t, a_t\right)$ is a function that represents the probability of transitioning to the next state $s_{t+1}$ when a specific action $a_t$ is taken from the current state $s_t$. In the DODE problem, this function describes how the traffic conditions change when a new OD matrix is input under given traffic conditions. This transition process is performed using microscopic traffic simulations. Therefore, the transition function is not explicitly known and is treated as a black-box function. This is the reason why a model-free reinforcement learning approach is necessary.

\subsubsection{Reward}
In this problem, the reward $r_t$ is defined by the difference between the simulated link flow vector $d^{\prime}{ }_k$ and ground-truth link flow vector $d_k$. Since the input interval is set to be short, our formulation corresponds to the case where $T>K$. The reward function can be written as:
\begin{equation}
\label{eq:eq3}
r_t =
\begin{cases}
0, & t \neq \max(\psi(k)), \\
-\left\|d'_k-d_k\right\|_2^2, & t=\max(\psi(k)).
\end{cases}
\end{equation}
subject to $d^{\prime}{ }_k \sim \mathcal{F}\left(a_{\psi(k)} \mid s_{\min (\psi(k))-1}\right)$ with a given state $s_{\min (\psi(k))-1} \in \mathcal{S}$ and $a_{\psi(k)} \in \mathcal{A}^{|\psi(k)|}$.

Note that, in our framework, the input interval and the output interval generally differ, so we introduce a mapping function $\mathcal{M}$. For example, when the input period is 5 seconds and the output period is 300 seconds, the mapping function is $\mathcal{M}=\psi(k=1)=\{t \in \mathbb{Z} \mid 1 \leq t \leq 60\}$. In this case, the reward $r_t$ is 0 except at $r_{60}$, when the output is generated. To align the reinforcement learning objective of maximizing cumulative reward with our aim of minimizing this discrepancy, we simply define the reward as the negative of the error.

\subsubsection{Discount factor}
The discount factor $\gamma$ is a value between 0 and 1 that determines the present value of future rewards. When $\gamma=0$, the agent exhibits myopic behavior, considering only immediate rewards. Conversely, when $\gamma \rightarrow 1$, the agent exhibits a far-sighted attribute, valuing future rewards as much as current rewards. In this paper, we use $\gamma=0.99$ to emphasize long-term rewards and promote stable learning.

\subsection{Deep Reinforcement Learning for DODE Problem}
The agent's behavior is described by a policy $\pi\left(a_t \mid s_t\right)$, which maps each state $s_t$ to a probability distribution over actions $a_t$. The goal of reinforcement learning is to find an optimal policy $\pi^*$ that maximizes the expected sum of discounted rewards, also known as the return. This objective can be formally written as:
\begin{equation}
\label{eq:eq4}
\pi^*=\arg \max _\pi J(\pi)=\arg \max _\pi \mathbb{E}_{\tau \sim \pi}\left[\sum_{t=1}^T \gamma^{t-1} r_t\right]
\end{equation}
where $\tau=\left(s_0, a_1, s_1, a_2, \ldots\right)$ is a trajectory generated by following policy $\pi$.

To solve this problem, we employ multi-binary PPO, an adaptation of the PPO algorithm featuring a factorized Bernoulli policy head. PPO is a prominent model-free DRL algorithm \citep{ref20}. This algorithm directly parameterizes and optimizes the policy $\pi_\theta\left(a_t \mid s_t\right)$, where $\theta$ represents the policy's parameters. PPO utilizes an actor-critic architecture. The actor, implemented as the policy network $\pi_\theta\left(a_t \mid s_t\right)$, selects actions. The critic, represented by a value network $V_\phi\left(s_t\right)$ with its parameters $\phi$, evaluates the quality of a state by estimating its expected return (the state-value).

We model the action as a multi-binary vector: the actor maps each state to an $n$-dimensional logits vector, applies a sigmoid to obtain per-component probabilities, and samples each OD pair via a Bernoulli distribution. This defines a factorized policy whose components are conditionally independent given the state. While one could treat $\{0,1\}^n$ as a single categorical space with $2^n$ joint actions, that approach incurs a combinatorial explosion in parameterization and exploration. By contrast, the proposed factorized Bernoulli head scales linearly with the number of OD pairs in terms of the policy's output dimension and parameter count, yielding superior scalability. Although factorization assumes conditional independence, a shared backbone can still consider implicit dependencies through shared features. The potential bias and performance implications of this factorized action-head approximation are empirically examined in \textbf{Appendix C} by comparing it with a joint categorical PPO head under independent, grouped, and fully coupled OD demand scenarios.

To guide the actor's learning, PPO uses an advantage estimate, $\hat{A}_t$, which quantifies how much better a given action $a_t$ is compared to the average action in the state $s_t$. We compute this advantage using GAE, which provides a low-variance estimate by balancing immediate rewards with long-term value predictions from the critic. For improved stability, the advantages are normalized per batch.

In PPO, the actor and critic are often trained jointly by optimizing a single, composite objective function. This objective is composed of three main components: the clipped surrogate objective for the policy ($L^{CLIP}$), an error term for the value function ($L^{VF}$), and an entropy bonus to encourage exploration. These are formulated as follows:
\begin{equation}
\label{eq:eq5}
L^{CLIP}(\theta)=\mathbb{E}_t\left[\min \left(r_t(\theta) \hat{A}_t, \operatorname{clip}\left(r_t(\theta), 1-\epsilon, 1+\epsilon\right) \hat{A}_t\right)\right]
\end{equation}
\begin{equation}
\label{eq:eq6}
L^{VF}(\phi)=\mathbb{E}_t\left[\left(V_\phi\left(s_t\right)-V_t^{\text {target }}\right)^2\right]
\end{equation}
\begin{equation}
\label{eq:eq7}
L^{PPO}(\theta, \phi)=-\mathbb{E}_t\left[L^{CLIP}(\theta)-c_1 L^{VF}(\phi)+c_2 \mathcal{H}\left[\pi_\theta\left(\cdot \mid s_t\right)\right]\right]
\end{equation}
where $\hat{A}_t$ is the advantage estimate, $r_t(\theta)$ is the probability ratio between the new and old policies, $\epsilon$ is the clipping range, $V_t{ }^{\text {target}}$ is the target value for the critic calculated as $V_t{ }^{\text{target}}=\hat{A}_t+V_\phi\left(s_t\right), \mathcal{H}[\pi]$ is the policy entropy, and $c_1$ and $c_2$ are the coefficients for the value function loss and the entropy bonus, respectively.

This objective function discourages excessive policy updates by penalizing probability ratios $r_t(\theta)$ that fall outside the $[1-\epsilon, 1+\epsilon]$ interval. When PPO is applied to the DODE problem, the actor network learns a policy for OD vehicle departure decisions $a_t$ based on the state $s_t$. The critic provides an evaluation of the states, and the resulting advantage function $\hat{A}_t$ offers a concrete basis for credit assignment. Based on this, PPO progressively improves the demand generation policy through stable updates, aiming to minimize the detector data error over the entire analysis period.

The choice of PPO is deliberate and central to addressing the challenges of the microscopic DODE problem. First, the clipped surrogate objective ($L^{CLIP}$) is critical for ensuring learning stability. Microscopic simulations are inherently stochastic, meaning the same action in a similar state can lead to different outcomes. The trust region imposed by the clipping mechanism prevents the policy from overreacting to this noise, resulting in more reliable and monotonic improvements, as observed in our experiments. Second, the use of GAE within the actor-critic framework directly tackles the long-horizon credit assignment problem. Since rewards are only provided every 5 minutes (based on aggregated link flows), while actions are taken every 5 seconds, GAE allows the agent to properly evaluate the long-term consequences of its fine-grained decisions, making it a highly suitable choice for this problem.

\subsection{Computational Setup}

All experiments were conducted on a workstation equipped with an AMD Ryzen Threadripper PRO 5975WX CPU (32 physical cores and 64 logical processors) and 128 GB of RAM. The proposed method collected simulation experiences using multiple independent SUMO environments. Each environment was executed in a separate Python subprocess, and the resulting experiences were synchronously aggregated for PPO updates. The toy experiment used 20 parallel simulator environments, whereas the case study used 30. This vectorized rollout architecture is well suited to PPO because trajectories from multiple independent stochastic environments can be incorporated into each policy update. The resulting diversity of experiences can support more stable policy updates while also reducing elapsed turnaround time through parallel simulation. In contrast, within each trial, the conventional methods as implemented in this study evaluated one candidate solution at a time using a single SUMO environment. Accordingly, the reported wall-clock times reflect the evaluated implementations and their different degrees of parallelism and should not be interpreted as aggregate computational cost.

\section{Toy Experiment}

\subsection{Experimental Settings}
\begin{figure}[htbp]
  \centering
  \includegraphics[width=1.0\linewidth]{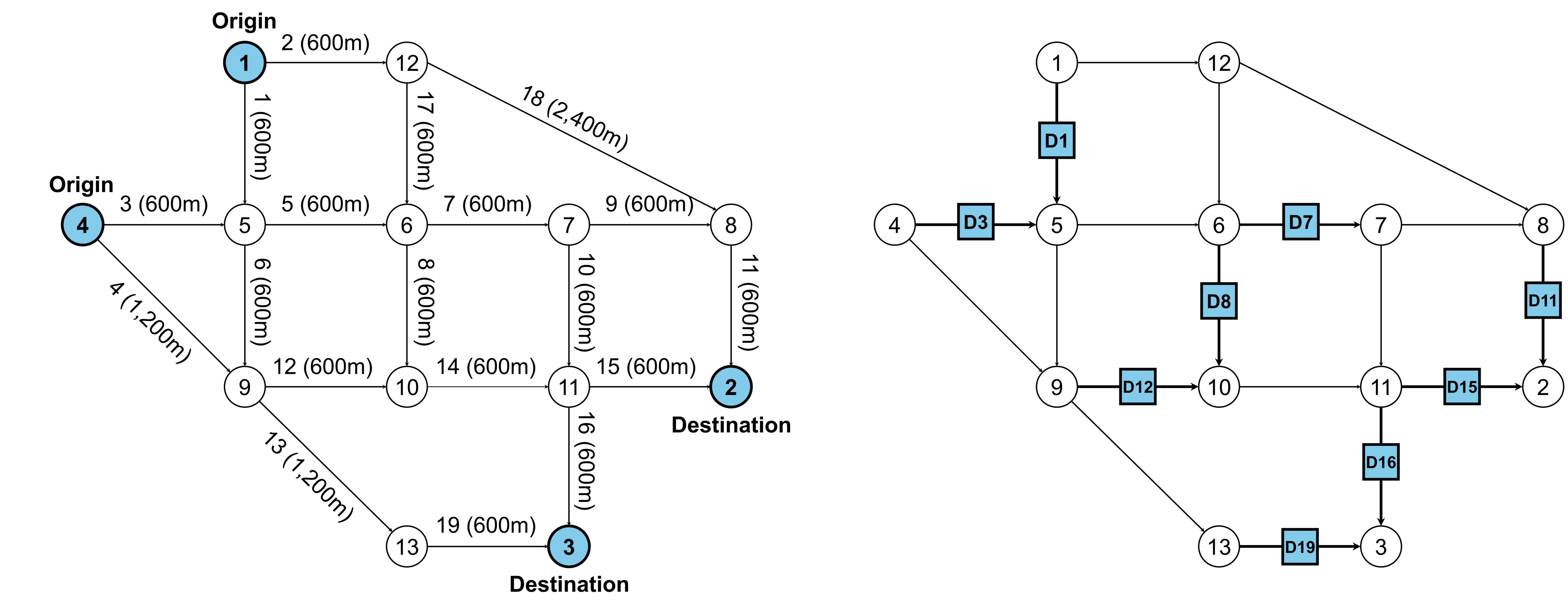}
  \caption{Nguyen and Dupuis network}
  \label{fig4}
\end{figure}

We constructed a toy experiment to evaluate the effectiveness of the proposed method. The Nguyen-Dupuis network, consisting of 4 OD pairs, 13 nodes, and 19 directed links, was adopted as a simple experimental setup to compare the effectiveness of the methodology \citep{ref47}. Each link is assumed to consist of a single lane. To ensure that the influence of the OD matrix is evident across a wide time range, we multiplied all link lengths by a factor of 3. We assume that virtual detectors are present on some of the links in the network shown in Figure~\ref{fig4}. These 9 detectors count the number of vehicles passing through each link every 5 minutes.

SUMO was used for microscopic traffic simulation \citep{ref48}. All hyperparameters are set to SUMO's default values. For example, all vehicles are implemented as passenger cars, and the Krauss model is used as the car-following model. Additionally, the acceleration is 2.6 m/s$^2$, and the deceleration is 4.5 m/s$^2$. Regarding the assignment, we update the time-dependent shortest path every 5 seconds for all vehicles and have them follow that path.

Ground-truth detector data is generated according to the following procedure. First, define the analysis period. In this study, we measure the flow of 9 links at 5-minute intervals for 30 minutes, resulting in a ground-truth dataset with 6 rows and 9 columns. Second, determine the total number of cars that will be input into the target network during the defined analysis period. To generate the ground-truth data, a total of 300 vehicles were created over the entire 30-minute analysis period. For each 5-second timestep, a departure decision was made probabilistically for each OD pair. As a result, the observed data that we need to fit through OD matrix estimation is shown in Table~\ref{tab:t2}.


\begin{table}[htbp]
  \centering
  \caption{Ground-truth detector data}
  \label{tab:t2}
  \begin{tabular*}{\linewidth}{@{\extracolsep{\fill}}lrrrrrrrrr}
    \toprule
      & \multicolumn{9}{c}{Link flow (veh/5 min)} \\
    \cmidrule(lr){2-10}
    Time (mm:ss) 
      & \multicolumn{1}{c}{D1} 
      & \multicolumn{1}{c}{D11} 
      & \multicolumn{1}{c}{D12} 
      & \multicolumn{1}{c}{D15} 
      & \multicolumn{1}{c}{D16} 
      & \multicolumn{1}{c}{D19} 
      & \multicolumn{1}{c}{D3} 
      & \multicolumn{1}{c}{D7} 
      & \multicolumn{1}{c}{D8} \\
    \midrule
    00:00--04:59 & 12 & 0  & 10 & 7  & 7  & 0  & 9  & 15 & 4 \\
    05:00--09:59 & 13 & 10 & 17 & 27 & 13 & 4  & 16 & 25 & 7 \\
    10:00--14:59 & 13 & 3  & 9  & 19 & 16 & 10 & 20 & 21 & 6 \\
    15:00--19:59 & 16 & 16 & 8  & 17 & 14 & 5  & 23 & 31 & 4 \\
    20:00--24:59 & 15 & 14 & 16 & 17 & 7  & 10 & 12 & 22 & 4 \\
    25:00--29:59 & 10 & 4  & 15 & 23 & 12 & 6  & 21 & 31 & 4 \\
    \bottomrule
  \end{tabular*}
\end{table}

In this study, the comparison targets are broadly classified into three categories. The first method reproduces detector data using different random seeds based on the assumed true demand. The second is a simultaneous optimization method. The third is a sequential optimization method. To demonstrate the limitations of existing methods, we applied the conventional input interval of 5 minutes and the proposed method's input interval of 5 seconds, respectively, and performed benchmarking. For the optimization algorithm, Bayesian optimization (BO), which exhibits eminent performance in optimizing high-cost objective functions, was applied \citep{ref49}. Additionally, we applied various black-box optimization techniques—including SPSA, differential evolution (DE), and covariance matrix adaptation evolution strategy (CMA-ES)—to the 5-minute interval simultaneous optimization and compared the results. Detailed explanations of each method are provided in Table~\ref{tab:t3}.


\begin{table}[!t]
  \centering
  \caption{Method description}
  \label{tab:t3}
  \setlength{\tabcolsep}{2.5pt}
  \renewcommand{\arraystretch}{1.12}

  {\footnotesize\hfill Note: $n$ is the number of OD pairs.\par}

  \vspace{0.5em}

  \begin{tabularx}{\linewidth}{
      @{}
      >{\raggedright\arraybackslash}p{0.07\linewidth}
      >{\raggedright\arraybackslash}p{0.17\linewidth}
      >{\raggedright\arraybackslash}p{0.20\linewidth}
      >{\raggedright\arraybackslash}p{0.12\linewidth}
      >{\raggedright\arraybackslash}p{0.13\linewidth}
      >{\raggedright\arraybackslash}X
      @{}
  }
    \toprule
    \multicolumn{6}{@{}l}{\textit{Methods}} \\
    \midrule
    Index & Abbreviation & Method & Algorithm & Iterations & Search space \\
    \midrule

    1 & True demand 
      & -- 
      & -- 
      & -- 
      & -- \\

    2 & RL-PPO 
      & DRL 
      & PPO 
      & 75 per environment 
      & $[0,1]^n$ \\

    3 & ST-BO (5min) 
      & \multirow{2}{*}{\begin{tabular}[c]{@{}l@{}}Simultaneous\\optimization\end{tabular}} 
      & \multirow{4}{*}{BO} 
      & \multirow{4}{*}{300} 
      & $[0,60]^n$ \\

    4 & ST-BO (5sec) 
      &  
      &  
      &  
      & $[0,1]^n$ \\

    5 & SQ-BO (5min) 
      & \multirow{2}{*}{\begin{tabular}[c]{@{}l@{}}Sequential\\optimization\end{tabular}} 
      &  
      &  
      & $[0,60]^n$ \\

    6 & SQ-BO (5sec) 
      &  
      &  
      &  
      & $[0,1]^n$ \\

    7 & ST-SPSA (5min) 
      & \multirow{3}{*}{\begin{tabular}[c]{@{}l@{}}Simultaneous\\optimization\end{tabular}} 
      & SPSA 
      & \multirow{3}{*}{600} 
      & \multirow{3}{*}{$[0,60]^n$} \\

    8 & ST-DE (5min) 
      &  
      & DE 
      &  
      &  \\

    9 & ST-CMA (5min) 
      &  
      & CMA-ES 
      &  
      &  \\

    \bottomrule
  \end{tabularx}

  \vspace{1em}

  \begin{tabularx}{\linewidth}{
      @{}
      >{\raggedright\arraybackslash}p{0.22\linewidth}
      >{\raggedright\arraybackslash}X
      >{\raggedright\arraybackslash}p{0.28\linewidth}
      @{}
  }
    \toprule
    \multicolumn{3}{@{}l}{\textit{Hyperparameters}} \\
    \midrule
    Algorithm & Hyperparameter & Value \\
    \midrule

    \multirow{5}{*}{PPO}
      & Learning rate & $3 \times 10^{-4}$ \\
      & Gamma & 0.99 \\
      & Number of steps & 180 \\
      & Lambda of GAE & 0.95 \\
      & Entropy coefficient & 0.01 \\

    \midrule
    \multirow{4}{*}{BO}
      & Number of initial points & 0 \\
      & Acquisition function & Expected improvement \\
      & Gaussian process kernel & Mat{\'e}rn ($\nu = 2.5$) \\
      & Gaussian process alpha & $1 \times 10^{-6}$ \\

    \midrule
    \multirow{4}{*}{SPSA}
      & Alpha & 0.602 \\
      & Gamma & 0.101 \\
      & Learning rate & 3.0 \\
      & Perturbation size & 2.0 \\

    \midrule
    \multirow{4}{*}{DE}
      & Population size & 24 \\
      & Differential weight & 0.8 \\
      & Dithering & $(0.5, 0.9)$ \\
      & Crossover rate & 0.9 \\

    \midrule
    \multirow{2}{*}{CMA-ES}
      & Population size & 16 \\
      & Sigma & 12.0 \\

    \bottomrule
  \end{tabularx}
\end{table}

\subsection{Results}
To evaluate the performance of the proposed approach, nine methods in Table~\ref{tab:t3} were compared. All methods were repeated 5 times in the identical setup. Since the research purpose is offline calibration, the points with the lowest error were primarily considered. We stored the OD matrices that achieved the maximum reward for each trial, then ran simulations with different seeds on the obtained OD matrices to fairly recalculate the average maximum reward by method.

\begin{figure}[htbp]
  \centering
  \includegraphics[width=1.0\linewidth]{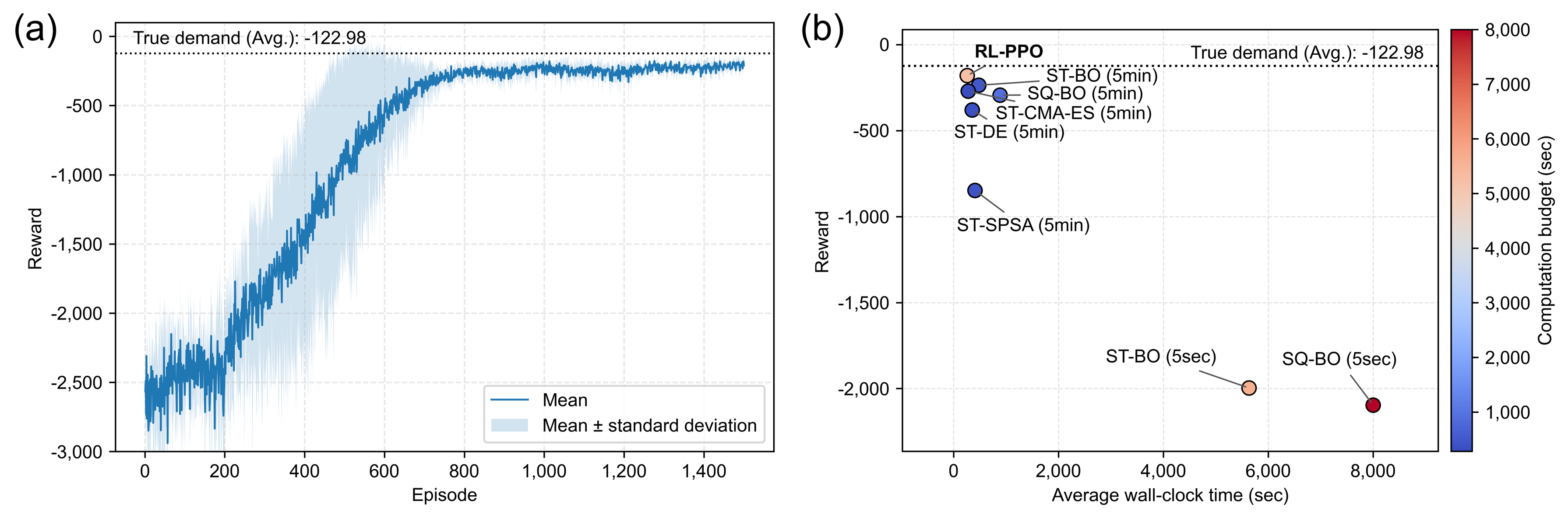}
  \caption{(a) Reward graph of the proposed method; (b) comparison of the highest reward, elapsed wall-clock time, and simulation budget across methods}
  \label{fig5}
\end{figure}

Figure~\ref{fig5}(a) displays the reward trend per episode of the proposed method. The bold line represents the average value, and the shaded area represents the range between the average value and the standard deviation. A dotted line shows the average maximum reward of the true demand scenario. The true demand scenario's maximum rewards are non-zero because of microscopic dynamics resulting from different random seeds, leading to different vehicle route outcomes.

Figure~\ref{fig5}(b) plots the average maximum reward, average wall-clock time, and computational budget for each method. Based on the average maximum reward, the results were as follows: True demand -122.98, RL-PPO -180.29, ST-BO (5min) -236.27, ST-BO (5sec) -1,996.84, SQ-BO (5min) -294.02, SQ-BO (5sec) -2,096.76, ST-SPSA (5min) -848.22, ST-DE (5min) -380.29, and ST-CMA-ES (5min) -270.44. This result indicates that our method achieves a 23.69\% improvement over the existing method with the best performance in reducing link-flow errors (-236.27 to -180.29). DRL requires numerous interactions with the environment, resulting in a high computational budget. However, when utilizing the PPO algorithm, multiple interactions can be performed in parallel within a single policy, leading to low wall-clock time. As a result, the proposed method achieved both minimal wall-clock time and maximum reward.

As emphasized in the early part of this paper, results reveal various limitations of existing methods. First, ST-BO (5sec) showed nearly the worst performance, contrasting with ST-BO (5min) showing the second-best. This highlights the complexity of the simultaneous optimization approach, where all decision variables are estimated simultaneously despite a 60-fold increase in dimension. Second, SQ-BO (5min) performed worse than ST-BO (5min), suggesting that, when the input dimension is small, exploring all decision variables simultaneously to avoid credit assignment is more appropriate than myopically estimating the OD matrix. Finally, SQ-BO (5sec) and ST-BO (5sec) show that reducing the temporal input unit is disadvantageous. Our method solves the DODE problem in microscopic traffic simulation by using an MDP formulation and DRL, while reducing the temporal input unit and achieving the best performance by adjusting demand at a higher resolution.

\begin{figure}[htbp]
  \centering
  \includegraphics[width=1.0\linewidth]{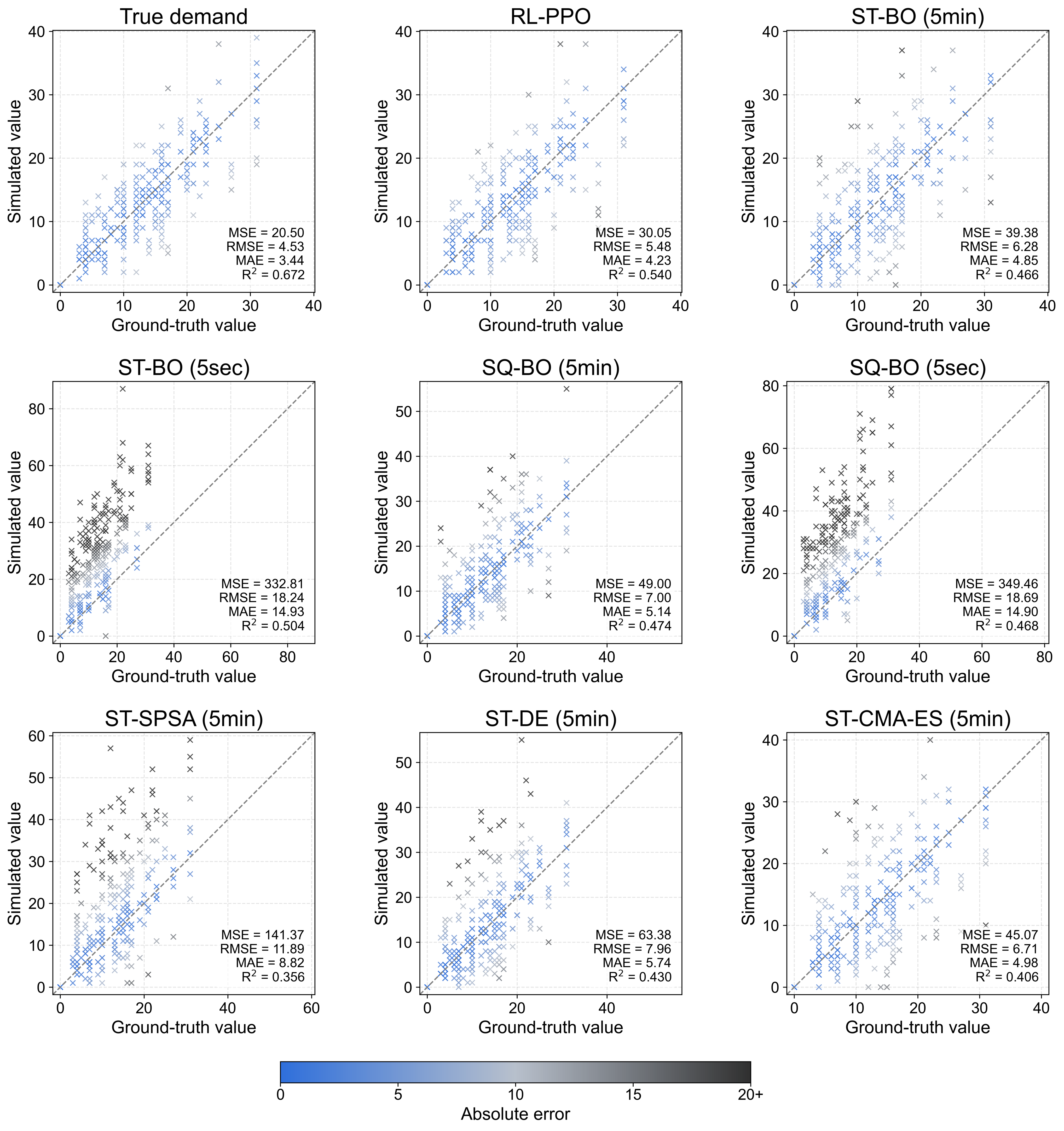}
  \caption{Scatter plots of detector data using estimated OD matrices by method}
  \label{fig6}
\end{figure}

Figure~\ref{fig6} presents scatter plots of data reproduced by the method. The x-axis of each subplot represents the ground-truth value, and the y-axis represents the simulated value.

\begin{figure}[htbp]
  \centering
  \includegraphics[width=1.0\linewidth]{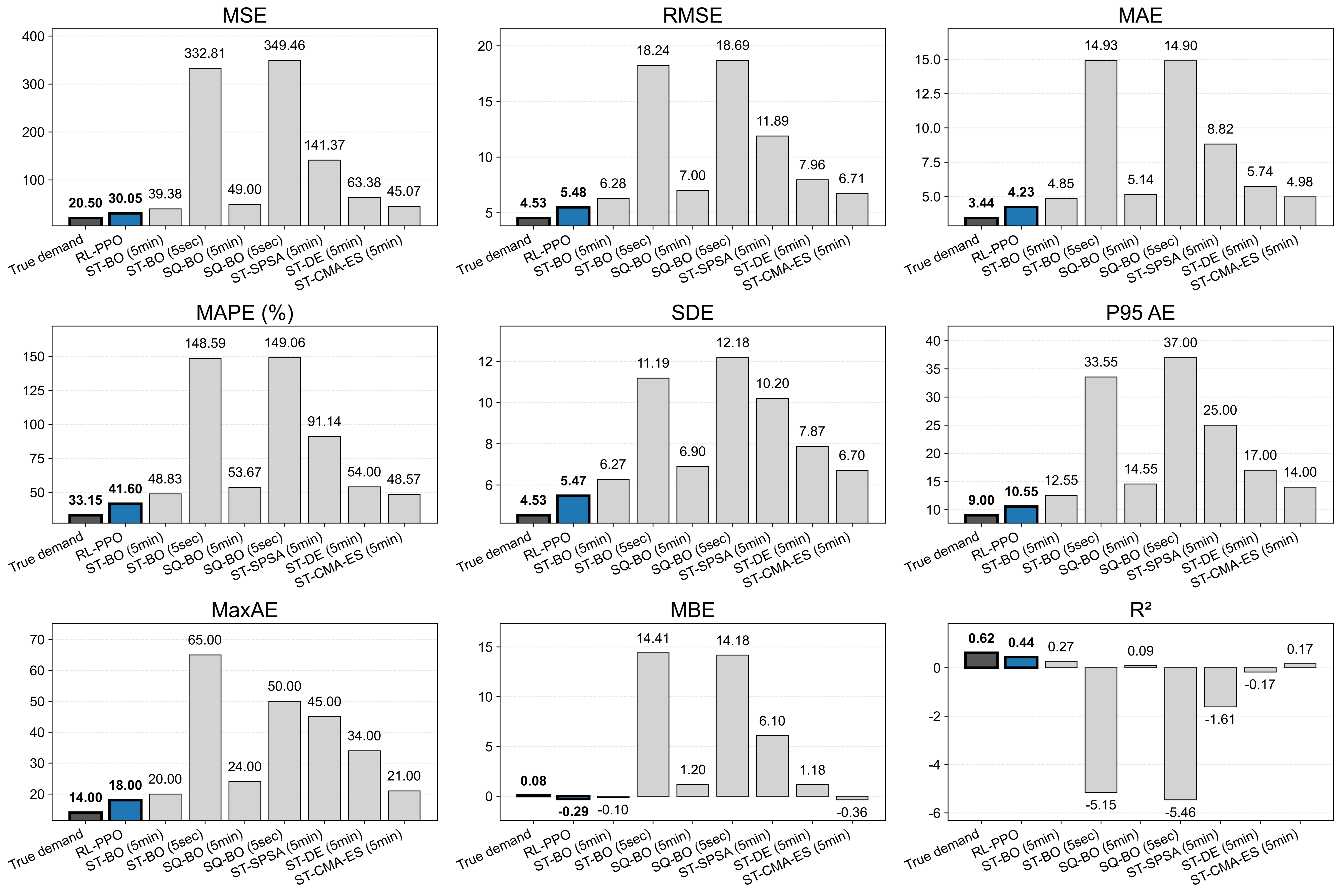}
  \caption{Comparison of multi-metric performance and stability across methods}
  \label{fig7}
\end{figure}

Figure~\ref{fig7} presents a set of performance and stability metrics, including MSE, root mean squared error (RMSE), mean absolute error (MAE), mean absolute percentage error (MAPE), standard deviation of error (SDE), 95th percentile of absolute error (P95 AE), maximum absolute error (MaxAE), mean bias error (MBE), and the coefficient of determination ($R^2$).

Together, these indicators assess the accuracy and stability of the reproduced link flows across all evaluation points in each subplot. MSE and RMSE represent the mean magnitude of squared errors and their square roots, respectively, while MAE measures the average absolute deviation between simulated and ground-truth flows. MAPE represents the relative error as a percentage and was computed excluding zero ground-truth points to ensure numerical stability. SDE quantifies the variability of the error distribution, P95 AE and MaxAE represent the high-percentile and maximum deviations, and MBE indicates whether the model systematically over- or underestimates link flows. Finally, $R^2$ measures how well the reproduced link flows explain the variation in the ground-truth data.

A notable observation is that microscopic traffic simulation is inherently stochastic, leading to considerable variations in link flows even under identical true-demand conditions (MSE = 20.50 $(\mathrm{veh}/5\mathrm{min})^2$, MAPE = 33.15\%). The proposed method achieves an MSE of 30.05 $(\mathrm{veh}/5\mathrm{min})^2$ and a MAPE of 41.60\%, closely matching the true-demand case. These results indicate that our method reproduces link-level dynamics with accuracy and stochastic stability comparable to the ground-truth demand (SDE = 5.47, $R^2$ = 0.44). In contrast, conventional methods such as ST-BO (5min) and SQ-BO (5min) show much higher errors (MSE = 39.38 and 49.00 $(\mathrm{veh}/5\mathrm{min})^2$, MAPE = 48.83\% and 53.67\%, respectively), indicating that their estimated OD matrices frequently over- or under-estimate link flows.

Although simultaneous optimization implicitly accounts for complex temporal and stochastic dependencies by jointly adjusting all timestep inputs, its computational complexity increases rapidly with the number of decision variables, reducing search efficiency and accuracy. Sequential optimization, while lowering the dimensionality of the problem, is inherently myopic and thus struggles to capture long-term temporal dependencies in stochastic environments. In contrast, the MDP-based DRL framework proposed in this study performs adaptive, stepwise decision-making while explicitly evaluating the long-term consequences of each action. Such a framework allows more effective credit assignment in highly stochastic and temporally correlated traffic systems. As a result, the proposed RL-PPO approach demonstrates robust accuracy and stability in microscopic environments compared to other methods.

\begin{figure}[htbp]
  \centering
  \includegraphics[width=1.0\linewidth]{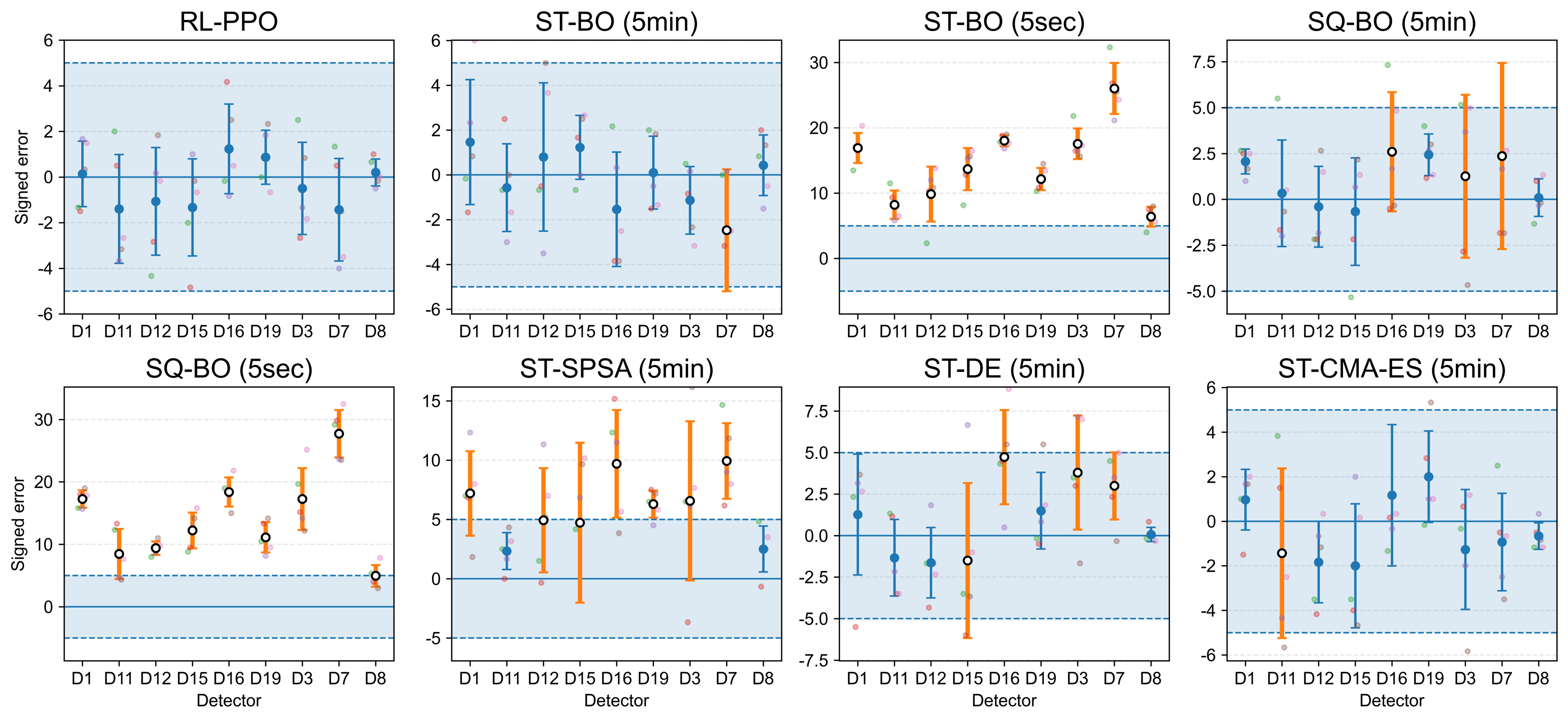}
  \caption{Detector-level TOST equivalence results for signed link-flow errors ($\alpha$= 0.05, $\Delta_d$= 5.0)}
  \label{fig8}
\end{figure}

We assessed detector-level practical equivalence between each calibrated simulation output and the reference simulation with true OD matrices using the two one-sided tests (TOST) procedure. Let $y_{r,t,d}$ denote the reference detector output and ${\hat{y}}_{r,t,d}$ denote the output from a candidate method, at time $t$, detector $d$, and trial $r$ (i.e., an independent run under a different random initialization). For each seed and detector, we computed the time-averaged signed error

\begin{equation}
\label{eq:eq8}
e_{r, d}=\frac{1}{T} \sum_{t=1}^T\left(\hat{y}_{r, t, d}-y_{r, t, d}\right),
\end{equation}

and tested whether the mean error $\mu_d=\mathbb{E}[e_{r,d}]$ lies within a pre-specified equivalence margin $\left[-\Delta_d,+\Delta_d\right]$. In our experiments, we used an absolute margin $\Delta_d$= 5.0 for all detectors and a significance level $\alpha$= 0.05. For each detector $d$, TOST was implemented via two one-sample, one-sided $t$-tests: (i) $H_{0,1}:\mu_d\le-\Delta_d$ vs.  $H_{1,1}:\mu_d>-\Delta_d$ and (ii) $H_{0,2}:\mu_d\geq+\Delta_d$ vs. $H_{1,2}:\mu_d<+\Delta_d$. A method was declared equivalent at detector $d$ only if both null hypotheses were rejected (i.e., both one-sided $p$-values were below $\alpha$). We additionally reported the 90\% confidence interval for $\mu_d$, consistent with TOST at $\alpha= 0.05$, and visualized it together with the equivalence band; per-trial error points were overlaid to illustrate run-to-run variability. As shown in Figure~\ref{fig8}, RL-PPO satisfied equivalence for all detectors, whereas each baseline method failed the equivalence test for at least one detector.

\subsection{Extensions for Regularization}

This section introduces an extension of the proposed method to address the multi-solution issue in the DODE problem. In actual OD matrix estimation tasks, obtaining the true OD demand values is generally challenging. Therefore, fit results for link flows are often used as an indirect indicator of estimation performance. However, since multiple solutions may exist that produce similar link flows, low error for link flow does not necessarily imply accurate OD matrix estimation, which is a limitation. This extension was evaluated only in the toy experiment, where the true OD matrix is available and the effect of regularization can be assessed directly in terms of OD recovery as well as link-flow fitting.

In our previous experiments, we assumed independence between OD matrices without any prior information. If we possess information about the temporal continuity of OD matrices or the ratio of demand for each OD pair, we can estimate a better OD matrix by adding a regularization term to impose constraints on the magnitude or direction of vector changes. In this case, weights can be set differently depending on the reliability of the information. Extension is achieved through two aspects: the reward function and the state representation. First, we accumulate the agent's actions in temporal blocks. The accumulated OD matrix $A_k$ within the $k$-th block is the sum of demands during that period, defined as follows using the notation from Table~\ref{tab:t1}:

\begin{equation}
\label{eq:eq9}
A_k=\sum_{t=\min(\psi(k))}^{\max(\psi(k))} a_t.
\end{equation}

Then, we add a magnitude penalty and a directional penalty to the reward function in Eq.~\eqref{eq:eq3} to penalize abrupt changes in the agent's behavior:

\begin{equation}
\label{eq:eq10}
r_t=-w_1 \Delta_{det}-w_2 \Delta_{OD}-w_3 \theta_{OD}
\end{equation}

where $t=\max (\psi(k))$, $\Delta_{det}=\left\|d^{\prime}{ }_k-d_k\right\|_2^2$, $\Delta_{O D}=\left\|A_k-A_{k-1}\right\|_1$, $\theta_{OD}=\arccos \left(\frac{A_k \cdot A_{k-1}}{\left\|A_k\right\|\left\|A_{k-1}\right\|}\right)$, $w_1=1.0$, $w_2=0.2$, $w_3=30.0$. $\Delta_{OD}$ encourages the agent to gradually increase or decrease demand, while $\theta_{OD}$ encourages the spatial distribution of OD demand to resemble that of the preceding block.

\begin{figure}[htbp]
  \centering
  \includegraphics[width=1.0\linewidth]{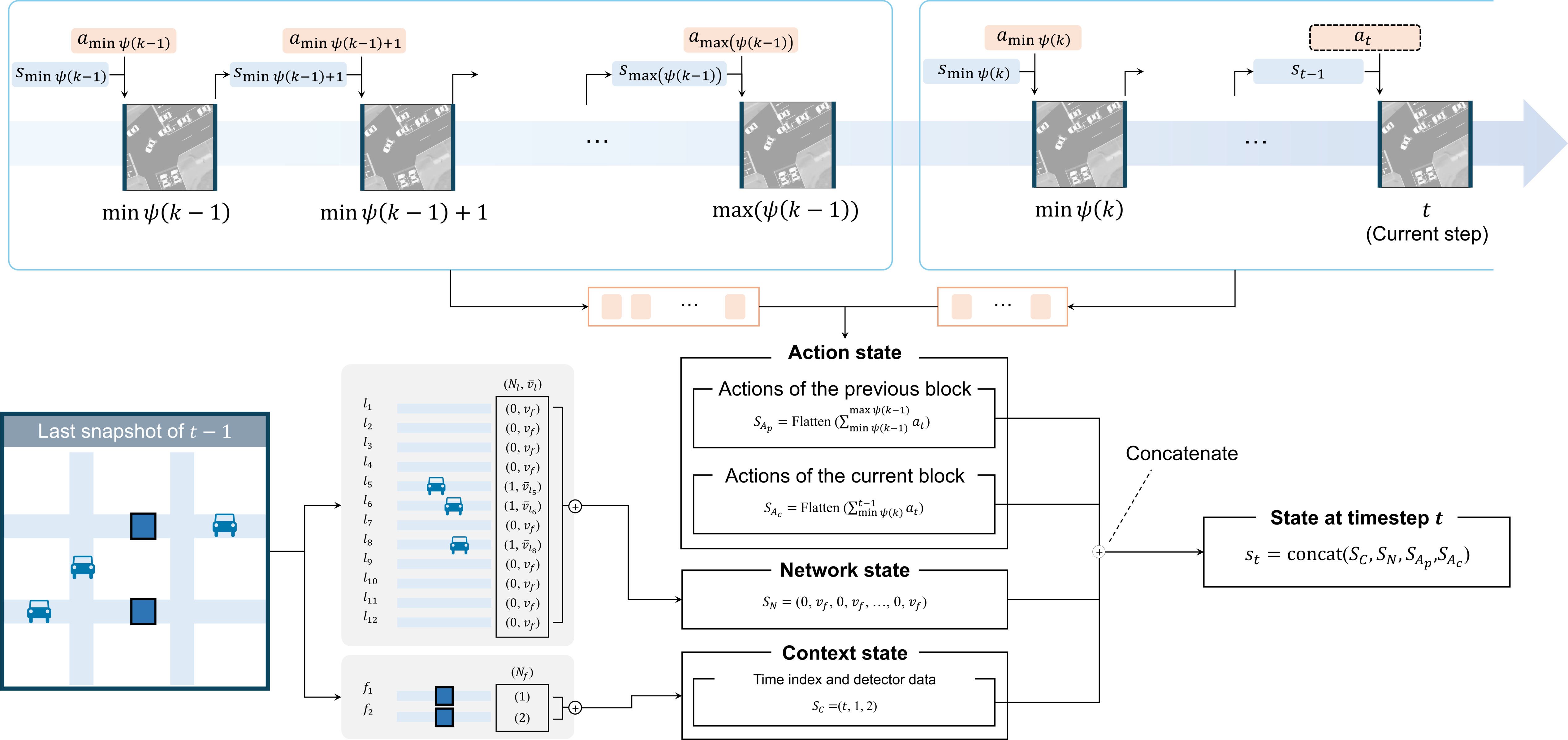}
  \caption{Extension of state representation}
  \label{fig9}
\end{figure}

Additionally, the state representation also needs to be modified. This is because the agent requires memory information about which actions were chosen in previous steps in order to improve rewards based on the changed reward function. As shown in Figure~\ref{fig9}, we included the cumulative OD matrix from the previous block and the cumulative OD matrix from the current block up to the present in the action state and provided it to the agent.

\begin{figure}[htbp]
  \centering
  \includegraphics[width=1.0\linewidth]{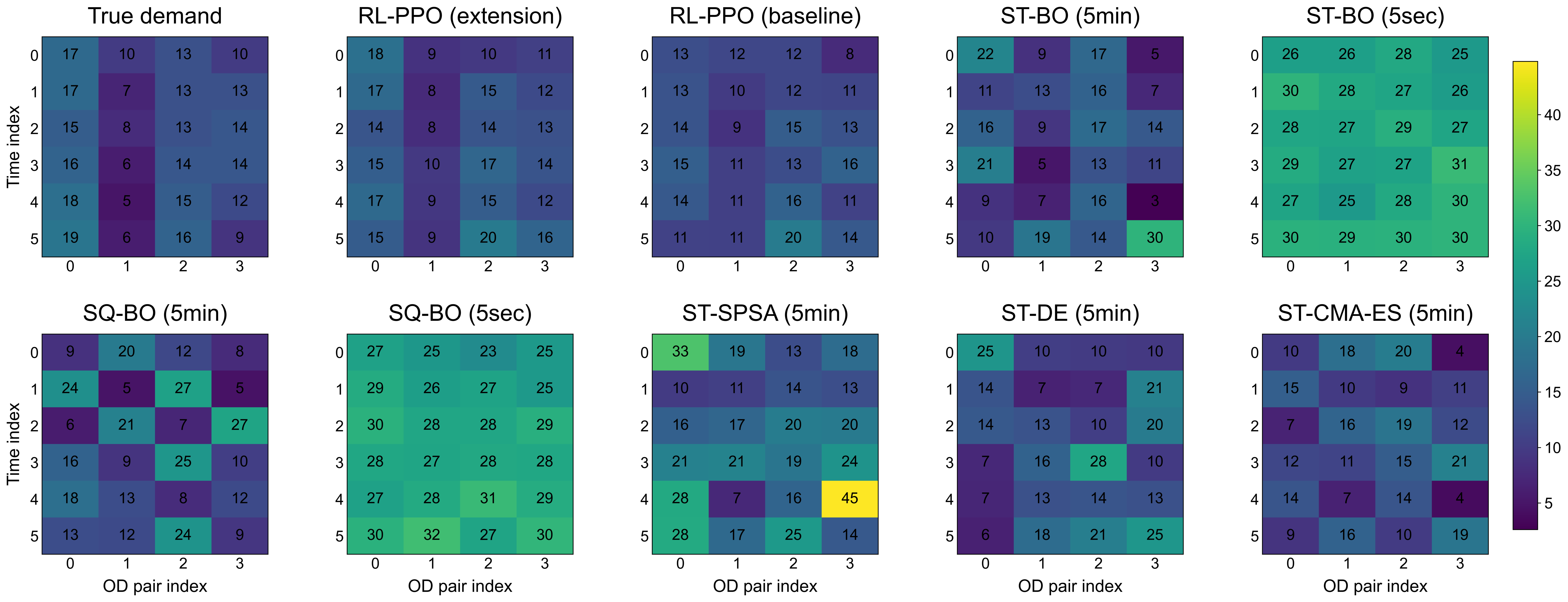}
  \caption{Heatmap showing the values of the actual OD matrix and the estimated OD matrix}
  \label{fig10}
\end{figure}

Figure~\ref{fig10} shows the estimated OD matrix results. The x-axis represents the OD pair index, and the y-axis represents the time index. The cell values are the average of the OD demands achieved when the maximum reward was attained for each method. Table~\ref{tab:t4} quantitatively shows how similar the true demand and estimated OD matrices are. The evaluation metrics used are MSE, which indicates absolute scale error; cosine similarity, which measures the agreement of the overall distribution ratio; mean structural similarity index measure (MSSIM) \citep{ref50}, which assesses the smoothness of local spatial structure; normalized Levenshtein distance for OD matrices (NLOD) \citep{ref51}, which simultaneously considers preference rankings and demand errors.

\begin{table}[htbp]
  \centering
  \caption{Quantitative similarity between the true and estimated OD matrices (MSE, cosine similarity, MSSIM, and NLOD)}
  \label{tab:t4}
  \setlength{\tabcolsep}{6pt}
  \renewcommand{\arraystretch}{1.12}
  \begin{tabular*}{\linewidth}{@{\extracolsep{\fill}}lrrrr@{}}
    \toprule
    Method 
      & \multicolumn{1}{c}{MSE} 
      & \multicolumn{1}{c}{Cosine similarity} 
      & \multicolumn{1}{c}{MSSIM} 
      & \multicolumn{1}{c}{NLOD} \\
    \midrule
    RL-PPO (extended) & 6.48   & 0.982 & 0.777 & 0.237 \\
    RL-PPO            & 11.14  & 0.967 & 0.661 & 0.313 \\
    ST-BO (5min)      & 45.57  & 0.884 & 0.432 & 0.399 \\
    ST-BO (5sec)      & 255.52 & 0.957 & 0.210 & 0.602 \\
    SQ-BO (5min)      & 55.94  & 0.882 & 0.245 & 0.527 \\
    SQ-BO (5sec)      & 252.81 & 0.951 & 0.200 & 0.549 \\
    ST-SPSA (5min)    & 105.40 & 0.926 & 0.219 & 0.466 \\
    ST-DE (5min)      & 57.16  & 0.871 & 0.063 & 0.492 \\
    ST-CMA (5min)     & 38.64  & 0.891 & 0.189 & 0.547 \\
    \bottomrule
  \end{tabular*}
\end{table}

The RL-PPO family generally recorded low MSE and high cosine similarity, reproducing the total volume and overall distribution patterns of the true OD matrix with relatively high fidelity. Notably, RL-PPO (extended) showed the smallest error with an MSE of 6.48, and its cosine similarity was also the highest at 0.982, demonstrating the best overall agreement in demand distribution ratios. From the perspective of structural similarity, the improvement of RL-PPO (extended) is also clearly observed. RL-PPO (extended) significantly enhanced local structural similarity with an MSSIM of 0.777 and also exhibited the smallest structural error based on destination preference with an NLOD of 0.237. In contrast, the baseline RL-PPO performed well in terms of MSE (11.14) but was relatively disadvantaged in MSSIM (0.661) and NLOD (0.313), suggesting that reducing scale error does not always lead to structural preservation.

\section{Case study}

\subsection{Experimental Settings}

\begin{figure}[htbp]
  \centering
  \includegraphics[width=1.0\linewidth]{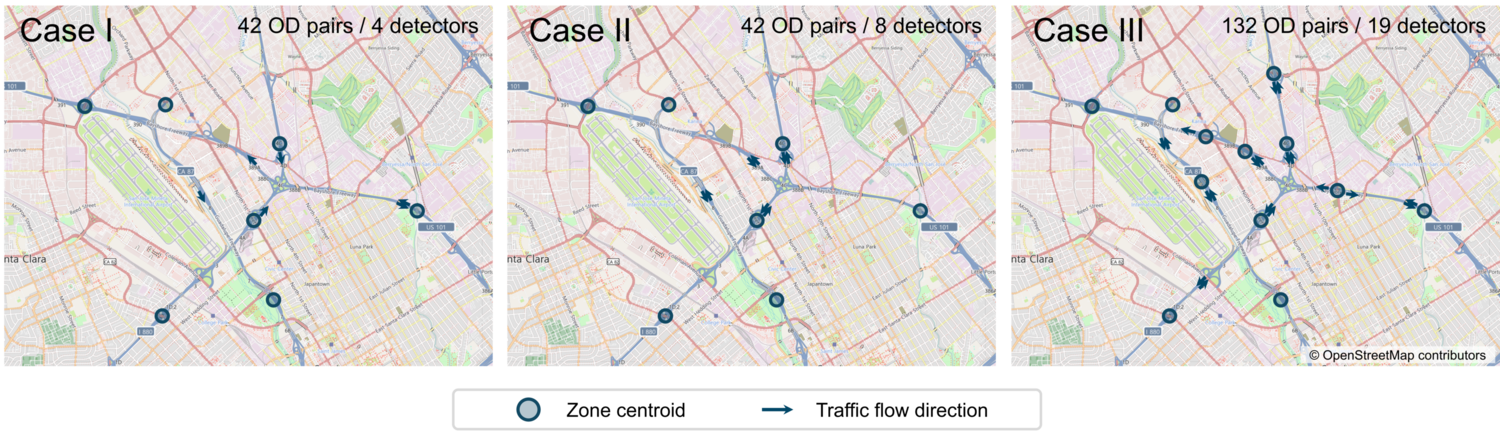}
  \caption{Experimental cases using the real-world network}
  \label{fig11}
\end{figure}

To evaluate the effectiveness of the proposed method in a real-world network, we conducted a case study. We selected a portion of the network in Santa Clara and San Jose, California, USA. This network is located where Interstate 880, US Route 101, and California State Route 87 intersect, and it has been utilized in the DODE problem benchmark \citep{ref41}. The latitude range for this area is (37.344, 37.396), and the longitude range is (-121.956, -121.870). The network consists of 245 nodes and 281 links. This network contains 240 OD pairs and 40 detectors, but we conducted experiments across three distinct cases to examine how the performance of the proposed method varies with the number of OD pairs or detectors. These cases are presented in Figure~\ref{fig11}.

We accessed data from the Caltrans performance measurement system (PeMS) for the period from 04:00 to 04:30 on January 26, 2026 \citep{ref52}. During this timeframe, a gradual increase in traffic volume from non-congested conditions to morning peak conditions can be observed. In applications, exact ground-truth OD demand matrices are generally unobservable. Therefore, following standard practice in OD matrix estimation studies, the field experiment evaluates the estimated OD demand inputs through their ability to reproduce observed detector link flows. This evaluation should be interpreted as observation-based field evaluation of OD demand estimation, not as proof of unique or exact OD matrix recovery. The estimated variables in our framework remain the time-dependent OD demands; link flows are used only as the observable consequences of these inputs through simulations.

PPO hyperparameters were adjusted according to the scale and temporal characteristics of each experimental setting. The toy network has a small OD dimension and relatively short credit-assignment horizon, so a standard PPO configuration with a moderate rollout length and entropy regularization was sufficient. In contrast, the case-study networks involve larger OD spaces, noisier detector responses, and longer delayed effects between OD actions and observed link flows. Therefore, we used longer rollouts, larger batches, a higher discount factor, and a more conservative learning rate to stabilize policy updates.

\subsection{Results}

\begin{figure}[htbp]
  \centering
  \includegraphics[width=1.0\linewidth]{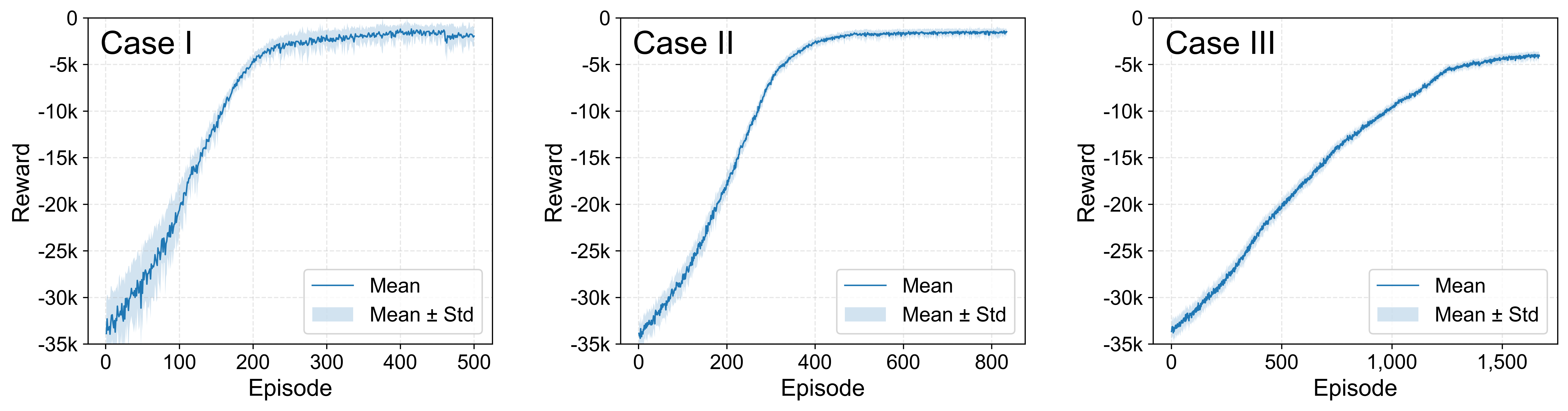}
  \caption{Reward graph of the proposed method by experimental case (Case study)}
  \label{fig12}
\end{figure}

The results were obtained by repeating the process five times with different random seeds for the estimated OD matrix. Figure~\ref{fig12} reports the reward graph of RL-PPO for the three cases, plotted as the mean $\pm$ standard deviation of episodic returns across 30 parallel environments. Across all cases, the policy exhibits stable improvement, while convergence becomes slower as the problem size increases. We benchmark RL-PPO against ST-BO (5min), which was the best-performing conventional baseline in the toy experiment. RL-PPO required 162.68 min (Case I), 255.56 min (Case II), and 927.09 min (Case III), whereas ST-BO (5min) required 183.47 min (Case I), 480.06 min (Case II), and 1,450.24 min (Case III).

\begin{figure}[htbp]
  \centering
  \includegraphics[width=1.0\linewidth]{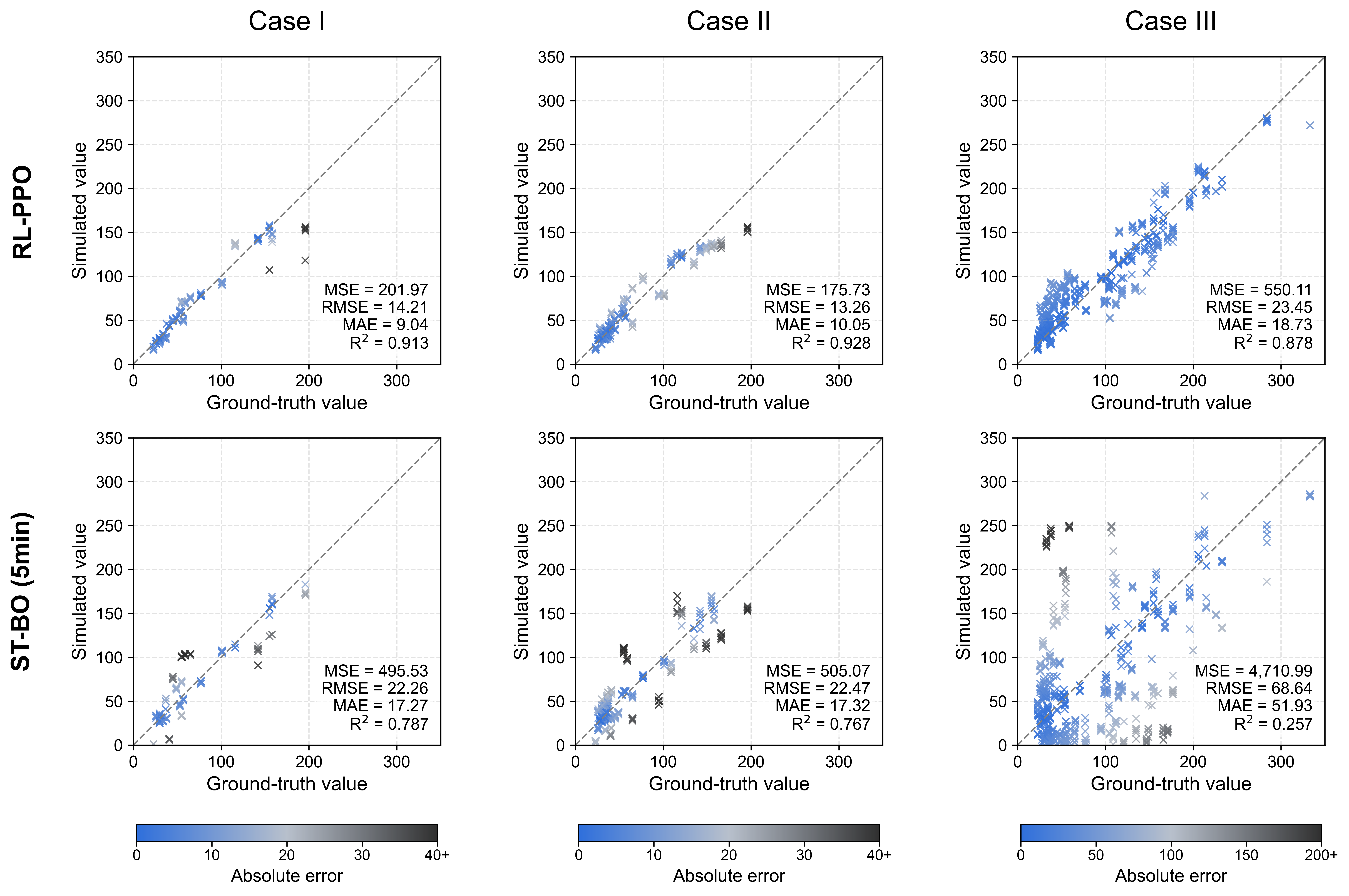}
  \caption{Scatter plots of detector data using estimated OD matrices by method (Case study)}
  \label{fig13}
\end{figure}

Figure~\ref{fig13} compares detector flow fitting between the two methods. Table~\ref{tab:t5} summarizes wall-clock time and error metrics. The error metrics reported in Table~\ref{tab:t5} are pooled point estimates calculated by aggregating the outputs from the five evaluation runs. Based on five paired evaluations using the same random seeds, the differences between RL-PPO and ST-BO were statistically significant for MSE, RMSE, MAE, and $R^2$ in all three cases (two-sided paired $t$-tests, all $p \leq 0.0124$). RL-PPO reduces MSE from 495.53 to 201.97 in Case I, from 505.07 to 175.73 in Case II, and from 4,710.99 to 550.11 in Case III, corresponding to 59.2\%, 65.2\%, and 88.3\% reductions, respectively. RL-PPO consistently demonstrated outstanding performance compared to the baseline, not only in MSE but also in RMSE, MAE, and $R^2$.

\begin{table}[!t]
  \centering
  \caption{Comparison of wall-clock time and calibration performance by case and method}
  \label{tab:t5}
  \setlength{\tabcolsep}{3pt}
  \renewcommand{\arraystretch}{1.15}

  \begin{tabularx}{\linewidth}{
      @{}
      >{\raggedright\arraybackslash}p{0.22\linewidth}
      >{\raggedright\arraybackslash}p{0.18\linewidth}
      >{\raggedleft\arraybackslash}p{0.17\linewidth}
      >{\raggedleft\arraybackslash}p{0.17\linewidth}
      >{\raggedleft\arraybackslash}p{0.17\linewidth}
      @{}
  }
    \toprule
    Category 
      & Subcategory 
      & \multicolumn{1}{c}{\shortstack{Case I\\(42 OD pairs,\\4 detectors)}}
      & \multicolumn{1}{c}{\shortstack{Case II\\(42 OD pairs,\\8 detectors)}}
      & \multicolumn{1}{c}{\shortstack{Case III\\(132 OD pairs,\\19 detectors)}} \\
    \midrule

    Wall-clock time (min)
      & RL-PPO 
      & 162.68 
      & 255.56 
      & 927.09 \\
    & ST-BO (5min) 
      & 183.47 
      & 480.06 
      & 1,450.24 \\

    \midrule
    MSE
      & RL-PPO\textsuperscript{$a$} 
      & 201.97 
      & 175.73 
      & 550.11 \\
    & ST-BO (5min)\textsuperscript{$b$} 
      & 495.53 
      & 505.07 
      & 4,710.99 \\

    \midrule
    \multirow{2}{*}{\begin{tabular}[c]{@{}l@{}}MSE per number\\of detectors\end{tabular}}
      & RL-PPO 
      & 50.49 
      & 21.97 
      & 28.95 \\
    & ST-BO (5min) 
      & 123.88 
      & 63.13 
      & 247.95 \\

    \midrule
    Relative reduction in MSE
      & $\text{value}=100(b-a)/b$
      & 59.2\% 
      & 65.2\% 
      & 88.3\% \\

    \midrule
    RMSE
      & RL-PPO 
      & 14.21 
      & 13.26 
      & 23.45 \\
    & ST-BO (5min) 
      & 22.26 
      & 22.47 
      & 68.64 \\

    \midrule
    MAE
      & RL-PPO 
      & 9.04 
      & 10.05 
      & 18.73 \\
    & ST-BO (5min) 
      & 17.27 
      & 17.32 
      & 51.93 \\

    \midrule
    $R^2$
      & RL-PPO 
      & 0.913 
      & 0.928 
      & 0.878 \\
    & ST-BO (5min) 
      & 0.787 
      & 0.767 
      & 0.257 \\

    \bottomrule
  \end{tabularx}
\end{table}

RL-PPO consistently outperforms ST-BO (5min) across all three cases, but the source of improvement differs by case. Comparing Case I and Case II isolates the effect of detector coverage because the OD dimension remains fixed at 42 pairs. Increasing the number of detectors from 4 to 8 improves the calibration fit: RL-PPO reduces the raw MSE from 201.97 to 175.73, decreases the normalized error from 50.49 to 21.97 per detector, and increases $R^2$ from 0.913 to 0.928. Although the wall-clock time increases from 162.68 to 255.56 min because the agent must match a larger set of detector-flow targets, RL-PPO remains substantially faster than ST-BO (5min), whose wall-clock time increases from 183.47 to 480.06 min. The relative MSE reduction over ST-BO also increases from 59.2\% to 65.2\%. These results suggest that additional detector coverage improves the identifiability of the calibration problem and that the proposed approach can exploit the additional observations more effectively than the baseline.

Comparing Case II and Case III further examines scalability when both the OD dimension and detector coverage increase. As expected, the absolute computational burden increases in the larger setting; however, the comparative advantage of RL-PPO becomes stronger rather than weaker. In Case III, which includes 132 OD pairs and 19 detectors, RL-PPO reduces MSE from 4,710.99 to 550.11, corresponding to an 88.3\% reduction, and improves $R^2$ from 0.257 to 0.878. It also reduces wall-clock time from 1,450.24 to 927.09 min relative to ST-BO (5min). This indicates that the proposed method becomes more advantageous as the calibration problem becomes more complex, because the learned policy can make state-dependent sequential decisions instead of repeatedly searching over a high-dimensional OD matrix as a decision vector. Nevertheless, the absolute training time in the largest case remains non-negligible, so further work on action-space structuring, OD clustering, hierarchical control, and reward decomposition is still necessary for deployment on larger urban networks.

\section{Conclusions}

This study introduced a reinforcement learning-based OD matrix estimation method to address the credit assignment problem, a persistent issue in DODE for microscopic traffic simulations. We address the limitations of conventional methods—namely, the complexity of simultaneous optimization and the myopia of sequential optimization—by redefining the DODE problem within an MDP framework and applying model-free DRL to derive optimal OD input sequences that consider long-term, stochastic effects.

Experimental results show that our method achieves significantly lower errors compared to existing simultaneous and sequential optimization approaches. In the toy network, compared to the baseline that showed the highest performance, our method improved the link-flow MSE by 23.7\%. A case study on a real-world freeway subnetwork using PeMS data indicates that the approach can extend beyond the toy experiment and improve link-flow calibration performance within the tested network scales. In three cases, our method reduced link-flow MSE by up to 88.3\% compared to the baseline and achieved lower wall-clock time while improving fitting accuracy.

The core contribution of this study is a novel reinforcement learning-based approach to the DODE problem that overcomes the trade-off between high-dimensionality and myopic decision-making inherent in existing optimization-based methodologies. The proposed MDP and DRL framework addresses the complexity and stochasticity of microscopic simulations by evaluating the long-term impacts of each action, thereby addressing the credit assignment problem in DODE. The effectiveness of this approach is supported by the use of multi-binary PPO, whose clipped surrogate objective helps stabilize policy updates in the presence of simulation noise and delayed rewards. Overall, the proposed framework provides a promising approach for high-resolution OD demand calibration in microscopic simulations and offers a foundation for future work on scalable calibration and operational applications.

Future research is recommended to focus on improving scalability and efficiency. Promising directions include learning concise state representations, structuring the action space through methods like OD clustering or hierarchical demand control, and adopting diversified or multi-objective reward shaping. Another direction is partial observability. In this experiment, we empirically validated the approximate MDP formulation using link counts and speeds; however, our system remains partially observable regarding the precise location, speed, and routes of individual vehicles. In this case, DODE is more naturally formulated as a POMDP, suggesting the need for memory and belief-state estimation. Integrating recurrent policies and explicit state-estimation modules that fuse multiple data sources is a promising path to improve robustness and transferability under realistic constraints.


\section*{Acknowledgments}
The authors used GPT-5.5 to review the code and manuscript, and they assume full responsibility for the final edited content.

\section*{Author Contributions}
The authors confirm contribution to the paper as follows: conceptualization: Min and Kim; data curation: Min; formal analysis: Min; methodology: Min, Choi, and Kim; writing---original draft: Min; writing---review and editing: Min, Choi, and Kim; draft manuscript preparation: Min, Choi, and Kim. All authors reviewed the results and approved the final version of the manuscript.

\section*{Data Availability}
The source code is available at \url{https://github.com/dgmin-kr/dode-rl-sumo}.

\section*{Declaration of Conflicting Interests}
The authors declared no potential conflicts of interest with respect to the research, authorship, and/or publication of this article.

\section*{Funding}
This work was supported by the Korea Institute of Police Technology (No. 092021C28S02000) and the National Research Foundation of Korea (No. 00409860). This work was also supported by the Advanced GPU Utilization Support Program funded by the Government of the Republic of Korea (Ministry of Science and ICT).


{\small
\bibliographystyle{unsrtnat}
\bibliography{references}
}


\clearpage
\setcounter{section}{1}
\setcounter{subsection}{0}
\renewcommand{\thesection}{\Alph{section}}
\renewcommand{\thesubsection}{\thesection.\arabic{subsection}}
\setcounter{figure}{0}
\renewcommand{\thefigure}{A.\arabic{figure}}
\setcounter{table}{0}
\renewcommand{\thetable}{A.\arabic{table}}
\setcounter{equation}{0}
\renewcommand{\theequation}{A.\arabic{equation}}
\section*{Appendix A}

This section outlines an experimental procedure to empirically support the question: ``Can observation states be treated as MDP states?'' The core idea is as follows:
\begin{itemize}
    \item If the formulation of this problem approximately satisfies the Markov property, providing additional past states and actions ($s_{t-1}$, $a_{t-1}$, ..., $s_{t-k}$, $a_{t-k}$) should not meaningfully improve the prediction performance for $s_{t+1}$ or $r_t$.
    \item We construct a dataset of random actions to exclude policy dependence and explore a state space.
    \item Using the same function approximator, we compare the one-step prediction error between the Markov feature model and the history-augmented model.
\end{itemize}

\subsection{Experimental Settings}
All experimental settings are identical to those in the main text. Behavior is determined by Bernoulli sampling for each OD pair, with $p=0.2$ indicating whether a vehicle departs. We ran 1,000 episodes (360,000 steps), with each episode consisting of ($s_t$, $a_t$, $s_{t+1}$, $r_t$) over 360 steps. Using this dataset, we construct the Markov feature and History-augmented feature as follows:

\begin{equation}
z_t^M=\left[s_t, a_t\right]
\label{eq:eqA1}
\end{equation}
\begin{equation}
z_t^H=\left[s_t, a_t, s_{t-1}, a_{t-1}, \ldots, s_{t-k}, a_{t-k}\right]
\label{eq:eqA2}
\end{equation}

We apply the same evaluation procedure for each feature. First, we randomly split the episode data into training (80\%) and testing (20\%) sets. Second, we z-score normalize the input features. Third, we fit a one-step predictor using a multi-layer perceptron (MLP) neural network. The MLP consists of two hidden layers with 256 units each and ReLU activations, and is trained using the Adam optimizer with early stopping based on validation MSE. Separate models are trained to predict the next state and the reward. Fourth, we calculate the RMSE on the held-out test set. Finally, we evaluate how much the state and reward prediction performance improves when using History-augmented features compared to Markov features.

\subsection{Results}
The results are shown in Table~\ref{tab:ta}. Overall, using History-augmented features did not improve prediction performance compared to using Markov features. Instead, incorporating historical information increased the state prediction error (e.g., a degradation ranging from 0.94\% to 4.58\% for $k\le$ 10, and up to 24.55\% for $k$ = 30). Reward prediction errors also showed negligible changes of less than 1.07\%. 

These results indicate that appending past states and actions does not provide meaningful additional predictive power, and likely introduces noise or overfitting to the function approximator. Consequently, the chosen observation mapping $s_t$ captures sufficient information to support approximate MDP treatment. These results support treating the interaction as an approximate MDP in this paper and motivate the use of standard MDP-based approach for our experiments. We note, however, that this empirical check does not prove exact Markovness and that stronger partial observability can arise in field settings with sparser measurements.


\begin{table}[!t]
  \centering
  \caption{Comparison of prediction RMSEs between Markov and history-augmented features}
  \label{tab:ta}
  \setlength{\tabcolsep}{3pt}
  \renewcommand{\arraystretch}{1.15}

  \begin{tabularx}{\linewidth}{
      @{}
      >{\raggedright\arraybackslash}p{0.18\linewidth}
      >{\raggedright\arraybackslash}p{0.27\linewidth}
      >{\raggedleft\arraybackslash}X
      >{\raggedleft\arraybackslash}X
      >{\raggedleft\arraybackslash}X
      >{\raggedleft\arraybackslash}X
      >{\raggedleft\arraybackslash}X
      @{}
  }
    \toprule
    Task 
      & Features 
      & \multicolumn{1}{c}{$k=1$}
      & \multicolumn{1}{c}{$k=2$}
      & \multicolumn{1}{c}{$k=5$}
      & \multicolumn{1}{c}{$k=10$}
      & \multicolumn{1}{c}{$k=30$} \\
    \midrule

    \multirow{3}{*}{\begin{tabular}[c]{@{}l@{}}State\\prediction\end{tabular}}
      & Markov 
      & 0.7875 
      & 0.7873 
      & 0.7910 
      & 0.8049 
      & 0.8307 \\

      & History-augmented 
      & 0.7970 
      & 0.7947 
      & 0.8159 
      & 0.8418 
      & 1.0346 \\

      & Relative change 
      & +1.21\% 
      & +0.94\% 
      & +3.15\% 
      & +4.58\% 
      & +24.55\% \\

    \midrule

    \multirow{3}{*}{\begin{tabular}[c]{@{}l@{}}Reward\\prediction\end{tabular}}
      & Markov 
      & 4.6710 
      & 4.6737 
      & 4.6951 
      & 4.7294 
      & 4.8695 \\

      & History-augmented 
      & 4.6721 
      & 4.6768 
      & 4.6963 
      & 4.7183 
      & 4.8172 \\

      & Relative change 
      & +0.02\% 
      & +0.07\% 
      & +0.03\% 
      & -0.23\% 
      & -1.07\% \\

    \bottomrule
  \end{tabularx}
\end{table}

\clearpage
\setcounter{section}{2}
\setcounter{subsection}{0}
\renewcommand{\thesection}{\Alph{section}}
\renewcommand{\thesubsection}{\thesection.\arabic{subsection}}
\setcounter{figure}{0}
\renewcommand{\thefigure}{B.\arabic{figure}}
\setcounter{table}{0}
\renewcommand{\thetable}{B.\arabic{table}}
\setcounter{equation}{0}
\renewcommand{\theequation}{B.\arabic{equation}}
\section*{Appendix B}

This section conducts comparative experiments based on the presence or absence of each component of the context state. There are three components: network state, time index, and detector data memory.

\subsection{Experimental Settings}
All experimental settings are identical to those described in the main text, with each case repeated five times. There are three comparison cases: using only network state, using network state and time index, and using network state, time index, and detector data memory together. To determine the proper state configuration, the average reward and mean standard deviation from the final 10\% of episodes during training are examined.

\subsection{Results}
\begin{figure}[htbp]
  \centering
  \includegraphics[width=1.0\linewidth]{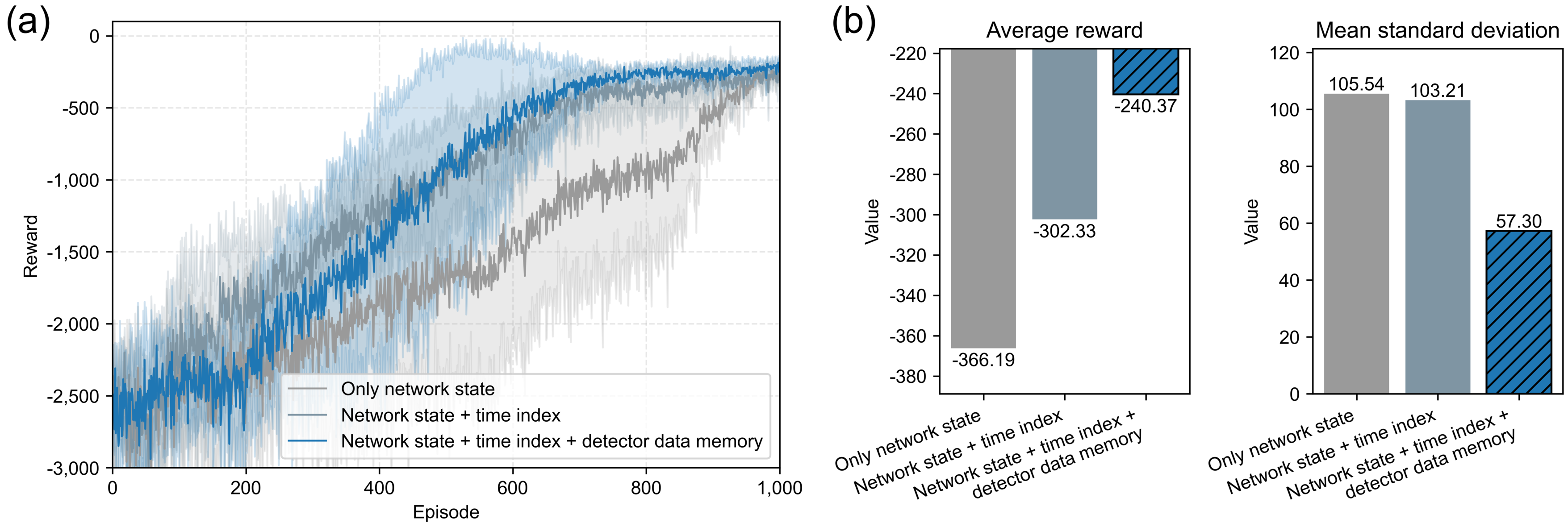}
  \caption{(a) Reward graph by state configuration; (b) comparison of the average reward and mean standard deviation by state configuration.}
  \label{figb}
\end{figure}

The experimental results are shown in Figure~\ref{figb}. When utilizing only the network state, the average reward was -366.19, and the mean standard deviation was 105.54. Next, when utilizing the time index together, the average reward was -302.33, and the mean standard deviation was 103.21. Finally, when all information was used together, the average reward was -240.37, with a standard deviation of 57.30. In summary, utilizing both the context state---comprising the time index and detector data memory---and the network state together enabled the most stable learning and yielded superior rewards.

\clearpage
\setcounter{section}{3}
\setcounter{subsection}{0}
\renewcommand{\thesection}{\Alph{section}}
\renewcommand{\thesubsection}{\thesection.\arabic{subsection}}
\setcounter{figure}{0}
\renewcommand{\thefigure}{C.\arabic{figure}}
\setcounter{table}{0}
\renewcommand{\thetable}{C.\arabic{table}}
\setcounter{equation}{0}
\renewcommand{\theequation}{C.\arabic{equation}}
\section*{Appendix C}

The proposed factorized Bernoulli policy head should be interpreted as a scalable policy parameterization rather than as a full joint distribution model over all OD departure patterns. With $n$ OD pairs, it requires only $n$ output probabilities, while a joint categorical policy requires $2^n$ action probabilities. This linear scaling is important for applying the proposed framework beyond the toy network. In the largest case study setting, the proposed head requires 132 output probabilities, whereas a joint categorical policy would require $2^{132}$ action probabilities. In addition, because all Bernoulli heads share the same state-dependent backbone, the policy can exploit common features such as network conditions, time, and detector data memory that influence multiple OD departure decisions. Thus, the factorized head can remain a practical approximation when OD dependencies are mainly reflected through the observed state.

However, the final action is still sampled separately for each OD pair. Therefore, it is necessary to examine whether this approximation creates bias when OD pairs are strongly coupled. In this section, we compare the proposed factorized Bernoulli PPO (FB-PPO) with a joint categorical PPO (JC-PPO) in the 4-OD toy network, where all $2^4=16$ joint departure patterns can still be explicitly represented. The joint categorical policy is used only as a diagnostic reference, not as a practical replacement for larger networks. The purpose of this appendix is to quantify how much OD-level bias appears under structurally coupled OD demand patterns, and whether this bias also degrades link-flow fitting performance.

\subsection{Experimental Settings}
All simulator settings, state representation, reward definition, and PPO hyperparameters follow the toy experiment in the main text, except that the demand-generation process is replaced by the diagnostic scenarios summarized in Table~\ref{tab:tc1} and the policy head is varied between FB-PPO and JC-PPO.


\begin{table}[htbp]
  \centering
  \caption{Diagnostic OD demand scenarios}
  \label{tab:tc1}
  \setlength{\tabcolsep}{4pt}
  \renewcommand{\arraystretch}{1.12}
  \begin{tabularx}{\linewidth}{
      @{}
      >{\raggedright\arraybackslash}p{0.24\linewidth}
      >{\raggedright\arraybackslash}X
      >{\raggedright\arraybackslash}p{0.34\linewidth}
      @{}
  }
    \toprule
    Scenario & Purpose & OD departure patterns \\
    \midrule
    Independent OD demand 
      & Reference case with independently generated OD-pair departures 
      & Independent Bernoulli departures for each OD pair \\

    Grouped OD demand 
      & Diagnostic case with mutually exclusive OD-group activation 
      & \begin{tabular}[t]{@{}l@{}}
          $[0,0,0,0]$, $[1,1,0,0]$, \\
          $[0,0,1,1]$
        \end{tabular} \\

    Fully coupled OD demand 
      & Diagnostic case with fully coupled OD-pair activation 
      & $[0,0,0,0]$, $[1,1,1,1]$ \\
    \bottomrule
  \end{tabularx}
\end{table}

The first method, denoted FB-PPO, is the proposed RL-PPO implementation with the shared-backbone factorized Bernoulli policy head used in the main experiments. The second method, denoted JC-PPO, uses a joint categorical policy head that selects one complete 4-bit OD departure pattern from the 16 possible patterns at each timestep. Each method was repeated over five trials. Then, we evaluate both link-flow fitting and OD recovery. Link-flow fitting is measured by link-flow MSE. OD recovery is measured by 5-minute block OD MSE, computed separately for each trial and then averaged over the five trials. To examine joint OD structure, we report total variation distance and the OD marginal error.

Link-flow MSE measures the mean squared error between reproduced and target counts, while OD MSE measures the mean squared error between the estimated and the true OD block matrix. Total variation distance quantifies the difference between the true and estimated empirical joint-action distributions. Marginal error measures the average error in origin productions and destination attractions, indicating how well the total outgoing and incoming OD demand is preserved.

\subsection{Results}

\begin{figure}[htbp]
  \centering
  \includegraphics[width=1.0\linewidth]{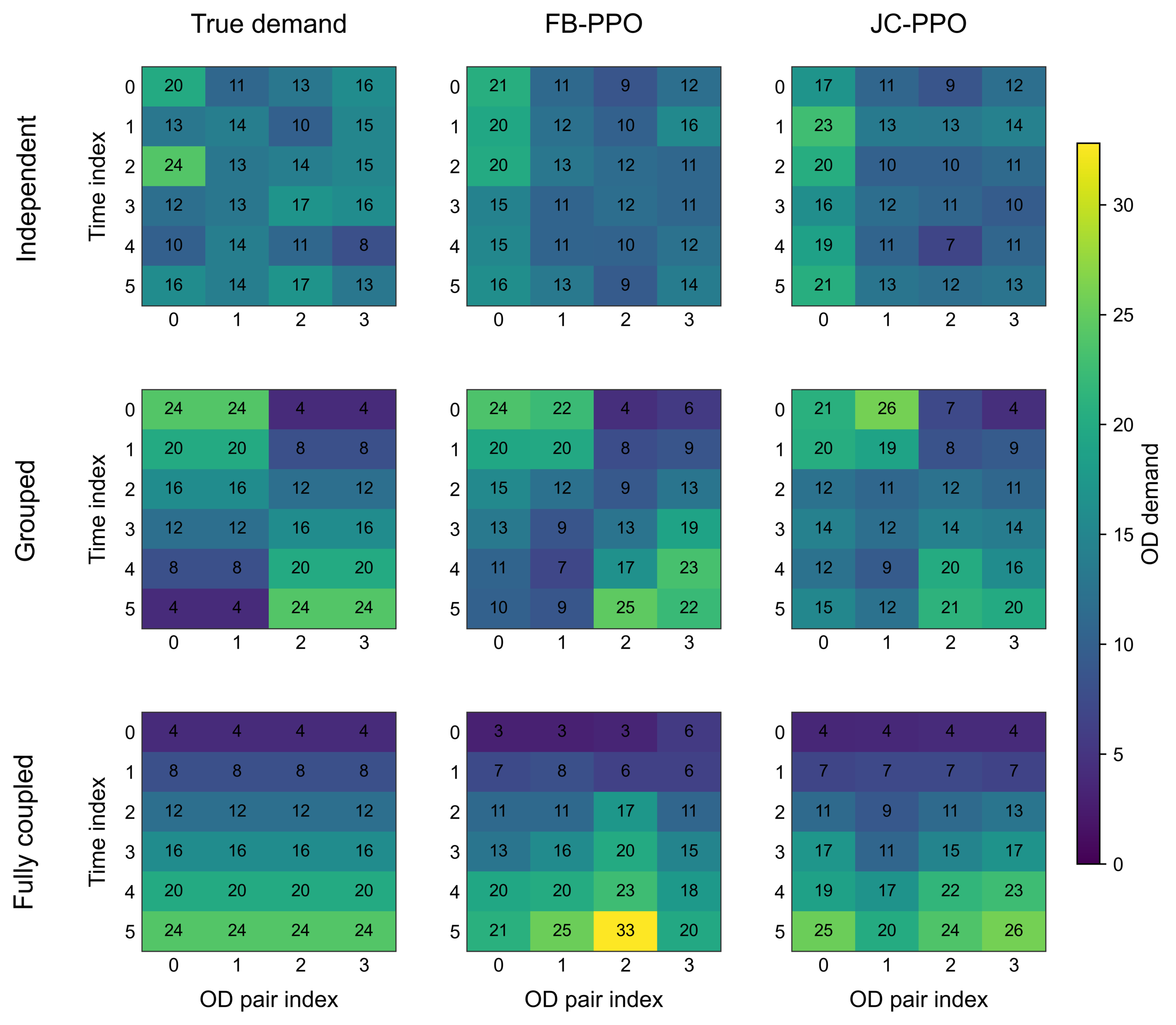}
  \caption{Comparison of true and estimated OD matrices under diagnostic demand scenarios}
  \label{figc}
\end{figure}

Table~\ref{tab:tc2} and Figure~\ref{figc} report the results of the action-head comparison. All values in this table are five-trial averages.


\begin{table}[htbp]
  \centering
  \caption{Results of the action-head comparison}
  \label{tab:tc2}
  \setlength{\tabcolsep}{3pt}
  \renewcommand{\arraystretch}{1.12}
  \begin{tabularx}{\linewidth}{
      @{}
      >{\raggedright\arraybackslash}p{0.24\linewidth}
      >{\raggedright\arraybackslash}p{0.10\linewidth}
      >{\raggedleft\arraybackslash}p{0.13\linewidth}
      >{\raggedleft\arraybackslash}p{0.09\linewidth}
      >{\raggedleft\arraybackslash}p{0.20\linewidth}
      >{\raggedleft\arraybackslash}p{0.13\linewidth}
      @{}
  }
    \toprule
    Scenario 
      & Method 
      & \multicolumn{1}{r}{\shortstack{Link-flow\\MSE}} 
      & \multicolumn{1}{r}{\shortstack{OD matrix\\MSE}} 
      & \multicolumn{1}{r}{\shortstack{Total variation\\distance}} 
      & \multicolumn{1}{r}{\shortstack{Marginal\\error}} \\
    \midrule
    Independent OD demand   & FB-PPO & 38.53 & \textbf{23.84} & \textbf{0.113} & \textbf{5.317} \\
    Independent OD demand   & JC-PPO & \textbf{35.47} & 28.90 & 0.233 & 6.000 \\
    Grouped OD demand       & FB-PPO & \textbf{25.79} & \textbf{22.52} & 0.589 & \textbf{4.617} \\
    Grouped OD demand       & JC-PPO & 26.09 & 23.86 & \textbf{0.547} & 5.283 \\
    Fully coupled OD demand & FB-PPO & \textbf{31.83} & \textbf{18.37} & 0.598 & 4.050 \\
    Fully coupled OD demand & JC-PPO & 31.89 & 21.88 & \textbf{0.501} & \textbf{3.783} \\
    \bottomrule
  \end{tabularx}
\end{table}

Relative to JC-PPO, FB-PPO changes the link-flow MSE by +8.6\%, -1.1\%, and -0.2\% in the independent, grouped, and fully coupled scenarios, respectively. Thus, the factorized Bernoulli policy head does not show systematic deterioration in link-flow fitting in these diagnostic cases. However, the joint-structure mismatch becomes more visible under coupled OD patterns: in the grouped and fully coupled scenarios, FB-PPO has larger total variation distances than JC-PPO (0.589 vs. 0.547 and 0.598 vs. 0.501, respectively), indicating that the factorized head is more prone to joint-structure mismatch even when link-flow fitting remains comparable.

In the independent OD demand scenario, the true demand is generated by independent Bernoulli departures for each OD pair, without imposed OD-group coupling. JC-PPO obtains a slightly lower link-flow MSE, whereas FB-PPO obtains lower OD MSE, lower total variation distance, and lower marginal error. This indicates that, when OD departures are generated independently, the factorized Bernoulli policy head is not disadvantaged by its structure.

In the grouped OD demand scenario, the true demand contains a stronger joint structure because two OD groups are activated as mutually exclusive patterns. FB-PPO achieves link-flow MSE and OD MSE comparable to JC-PPO. However, JC-PPO obtains lower total variation distance, indicating that the joint categorical head better preserves the empirical joint-action distribution when the true demand is supported by a small set of coupled OD patterns.

In the fully coupled OD demand scenario, the structural coupling is strongest because all OD pairs are activated together. Link-flow MSE is nearly identical between the two methods, and FB-PPO obtains lower OD MSE. However, JC-PPO achieves lower total variation distance and lower marginal error. This indicates that the joint categorical head better captures the true joint OD support, while the factorized Bernoulli policy head can still fit link flows and block-level OD volumes reasonably well.

Overall, JC-PPO is better suited for preserving empirical joint-action support when OD departures are strongly coupled. However, this advantage does not consistently translate into lower link-flow MSE or lower block-level OD MSE. These results suggest that the factorized Bernoulli policy head is a scalable approximation that remains suitable for DODE, but it can bias the recovered joint OD support when residual coupling among OD pairs remains strong after conditioning on the state.


\newpage

\end{document}